\documentclass[letterpaper, 10 pt, journal, twoside]{IEEEtran}
\usepackage{times}

\usepackage{graphics}           
\usepackage{times}              
\usepackage{amsmath}            
\usepackage{amssymb}            
\usepackage{graphicx}
\usepackage{algorithm}
\usepackage[noend]{algpseudocode}
\usepackage{booktabs}
\usepackage{color}
\usepackage[nocompress]{cite} 
\definecolor{instructioncolor}{rgb}{.5,.5,.5}

\usepackage[font=small]{caption}

\def\secref#1{Sec.~\ref{#1}}
\def\figref#1{Fig.~\ref{#1}}
\def\tabref#1{Tab.~\ref{#1}}
\def\eqref#1{Eq.~(\ref{#1})}

\makeatletter
\usepackage{xspace}
\DeclareRobustCommand\onedot{\futurelet\@let@token\@onedot}
\def\@onedot{\ifx\@let@token.\else.\null\fi\xspace}
 
\def\ie{i.e\onedot} 
\def\cf{cf\onedot}

\def\etal{{et al}\onedot}
\makeatother

\usepackage{array}
\newcolumntype{L}[1]{>{\raggedright\let\newline\\\arraybackslash\hspace{0pt}}m{#1}}
\newcolumntype{C}[1]{>{\centering\let\newline\\\arraybackslash\hspace{0pt}}m{#1}}
\newcolumntype{R}[1]{>{\raggedleft\let\newline\\\arraybackslash\hspace{0pt}}m{#1}}

\newcommand{\RR}{\mathbb{R}}

\newcommand{\ZZ}{\mathbb{Z}}

\newcommand{\mybold}[1]{\mbox{\boldmath$#1$}}

\newcommand{\vv}[1]{{\mybold #1}} 

\newcommand{\m}[1]{{\mbox{{\sffamily\slshape{#1\/}}}}}

\newcommand{\mq}[1]{{\mbox{{\sffamily{#1}}}}}

\newcommand{\tr}[0]{\sf T}              

\usepackage{booktabs}
\usepackage{siunitx}
\usepackage{multirow}
\usepackage{pifont}
\usepackage{amssymb}
\usepackage{amsmath}
\usepackage{subfigure}
\usepackage{tabularx}
\usepackage{comment}
\usepackage{mathtools} 
\usepackage{stmaryrd}
\usepackage{tabularx} 
\usepackage{color}
\usepackage{colortbl}
\usepackage[dvipsnames]{xcolor}
\usepackage[bookmarks=false]{hyperref}  

\newcommand{\cmark}{\ding{51}}%
\newcommand{\xmark}{\ding{55}}%
\renewcommand{\and}{\hspace{1.0cm}} 

\begin{document}

\title{PIN-SLAM: LiDAR SLAM Using a Point-Based Implicit Neural Representation for Achieving Global Map Consistency}

\author{Yue Pan, Xingguang Zhong, Louis Wiesmann, Thorbjörn Posewsky, Jens Behley, and Cyrill Stachniss}

\twocolumn[{%
\renewcommand\twocolumn[1][]{#1}%

\maketitle

\begin{center}
  \centering
  \vspace{-17pt}
  \captionsetup{type=figure}
  \includegraphics[width=0.97\linewidth]{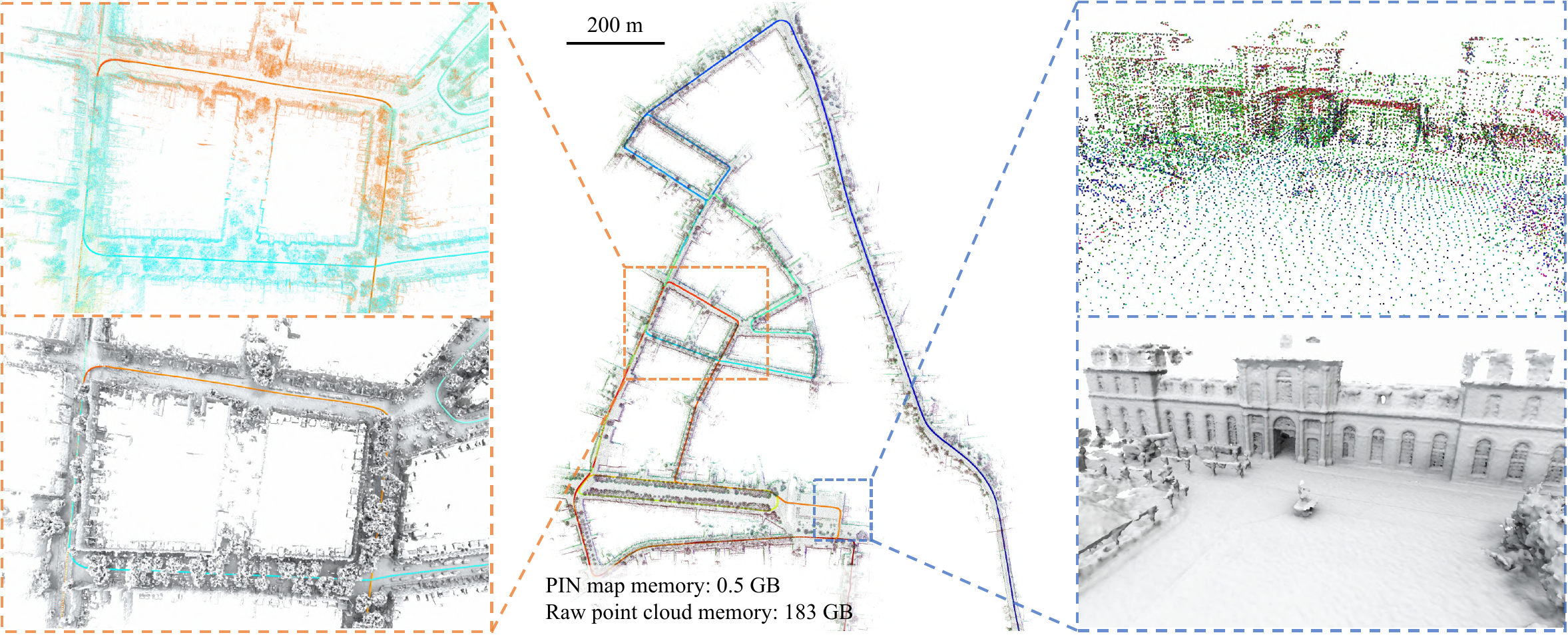}
  \setlength{\abovecaptionskip}{6pt}
  \captionof{figure}{We present PIN-SLAM, a novel LiDAR SLAM system using an elastic point-based implicit neural map representation. 
  Depicted in the middle, we show a large-scale globally consistent neural point map built with our approach using about 20,000 LiDAR scans recorded with a car without using any information from a GNSS, IMU or wheel odometry. 
  We can query the SDF value at an arbitrary position from the neural point map and reconstruct surface meshes. 
  The point colors represent the neural point feature after online optimization.
  On the left, we show the consistent neural points (top) and mesh (bottom) of a region traversed by the car multiple times indicated by the dashed orange box. 
  The colors of the neural points (top) represent timesteps when the point was added to the map. 
  On the right, we show the high-fidelity mesh (bottom) of a building reconstructed from the neural point map (top) of the region indicated by a dashed blue box.}
  \label{fig:mot}
  \vspace{0pt}
\end{center}%
}]

\makeatletter{\renewcommand*{\@makefnmark}{}
\footnotetext{Manuscript received: January 16, 2024; Revised: May 28, 2024; Accepted: June 26, 2024.}
\makeatother
\footnotetext{All authors are with the University of Bonn, Germany. Cyrill Stachniss is additionally with the Department of Engineering Science at the University of Oxford, UK, and with the Lamarr Institute for Machine Learning and Artificial Intelligence, Germany.}\makeatother
\footnotetext{This work has partially been funded by the European Union under the grant agreements No~101070405~(DigiForest), and
  by the Deutsche Forschungsgemeinschaft (DFG, German Research Foundation) under Germany's Excellence Strategy, EXC-2070 -- 390732324 -- PhenoRob, under STA~1051/5-1 within the FOR 5351~--~459376902~(AID4Crops).}\makeatother
}

\markboth{IEEE Transactions on Robotics. Preprint Version. Accepted June, 2024}{Pan \MakeLowercase{\textit{et al.}}: PIN-SLAM: LiDAR SLAM Using a Point-Based Implicit Neural Representation for Achieving Global Map Consistency}

\begin{abstract}
  
  Accurate and robust localization and mapping are essential components for most autonomous robots. 
  In this paper, we propose a SLAM system for building globally consistent maps, called PIN-SLAM, that is based on an elastic and compact point-based implicit neural map representation.
  Taking range measurements as input, our approach alternates between incremental learning of the local implicit signed distance field and the pose estimation using a correspondence-free, point-to-implicit model registration to the current local map. 
  Our implicit map is based on sparse optimizable neural points, which are inherently elastic and deformable with the global pose adjustment when closing a loop. Loops are also detected using the neural point features.
  Extensive experiments validate that PIN-SLAM is robust to various environments and versatile to different range sensors such as LiDAR and RGB-D cameras. 
  PIN-SLAM achieves pose estimation accuracy better or on par with the state-of-the-art LiDAR odometry or SLAM systems and outperforms the recent neural implicit SLAM approaches while maintaining a more consistent, and highly compact implicit map that can be reconstructed as accurate and complete meshes. 
  Finally, thanks to the voxel hashing for efficient neural points indexing and the fast implicit map-based registration without closest point association, PIN-SLAM can run at the sensor frame rate on a moderate GPU.

\end{abstract}

\section{Introduction}
\label{sec:intro}



\IEEEPARstart{S}{imultaneous} localization and mapping~(SLAM) based on range sensors, such as light detection and ranging~(LiDAR) sensors or RGB-D cameras, is a fundamental building block for autonomous mobile robots~\cite{grisetti2007tro,newcombe2011ismar,zhang2014rss,vizzo2023ral}. 
Various map representations~\cite{deschaud2018icra, behley2018rss, vizzo2021icra} and scan registration algorithms~\cite{besl1992pami, rusinkiewicz2001dim,pan2021icra-mvls} have been proposed over the last decades for efficient and high-fidelity representations of the environment as well as robust and accurate localization of the robot within the mapped environment.


The recent advance in neural implicit representation has shown several advantages over the classical explicit map representation widely used nowadays.
Instead of explicitly storing properties in a grid map, one can train a neural network to fit the observations in a scene. 
This allows for querying properties such as signed distances~\cite{park2019cvpr}, occupancy probability~\cite{mescheder2019cvpr}, colors~\cite{mildenhall2020eccv}, intensities~\cite{huang2023iccv}, and semantics~\cite{zhi2021iccv} at an arbitrary location within the scene implicitly using the neural network.
As a memory-efficient continuous representation, these kinds of implicit neural maps support efficient rendering~\cite{mueller2022acmgraphics}, surface reconstruction~\cite{zhong2023icra}, collision avoidance~\cite{ortiz2022rss} and path planning~\cite{yan2023iccv}. 
To enable the model to have a high representation capacity, current approaches often learn the neural implicit representation using optimizable local latent features instead of using a single globally shared neural network~\cite{sucar2021iccv,ortiz2022rss}.
The local latent features can be structured by a regular 3D grid~\cite{ mueller2022acmgraphics}, an axis-aligned tri-plane~\cite{peng2020eecv} with 2D grids, or a set of irregular points~\cite{xu2022cvpr}.
Consequently, several SLAM systems based on neural implicit representation have been proposed, mainly for RGB-D cameras operating indoor~\cite{sucar2021iccv, zhu2022cvpr,yang2022ismar, johari2023cvpr, sandstrom2023iccv} but also for LiDAR sensors operating outdoor~\cite{deng2023iccv}.
However, these SLAM approaches lack support for direct loop closure corrections and are not able to build globally consistent maps of larger scenes. This is mainly due to the usage of regular grid-based local feature embeddings, which are not elastic and resilient to loop corrections.
%

In this paper, we investigate the problem of realizing a SLAM system using an implicit neural map representation that supports globally consistent mapping.
%
We opt to use a neural point-based implicit representation, which has two main advantages over grid-based representations: the flexibility of spatial distribution and the elasticity for transformations.
Recent work~\cite{sandstrom2023iccv} only makes use of the first advantage for neural RGB-D SLAM at the cost of scalability and inefficient neighborhood querying. 
Instead, we exploit the second and more important advantage to build a globally consistent map that can be corrected while online mapping after closing a loop, which is essential for large-scale LiDAR-based SLAM. 
Our approach alternates between mapping, \ie, online learning of the local implicit signed distance map given poses, and odometry, \ie, tracking the next scan's pose given the current local map, additionally with the ability to correct the drift and keep a globally consistent map after closing a detected loop. 

For the mapping part, we adapt the training data sampling strategy and loss function used in our previous work~\cite{zhong2023icra} but replace the octree-based multi-resolution feature grid with a set of neural feature points. 
The signed distance function~(SDF) and additional properties like color or semantics at a query position are predicted by interpolating among the surrounding neural points. 
For each neural point, we concatenate its optimizable latent feature with the query position's coordinates in the neural point's coordinate system.
%
%
This concatenated feature is then decoded by a globally shared multi-layer perceptron (MLP) as the prediction for that neural point.
For efficient and scalable neighborhood querying for the interpolation, we use a voxel hashing data structure to index the neural points by keeping no more than one active neural point in each voxel. 
This enables our system to run at the sensor frame rate (10\,Hz) regardless of the scale of the mapped scene. 
To tackle the problem of ``catastrophic forgetting" in incremental learning, instead of using the less stable regularization-based strategy~\cite{zhong2023icra}, we use a local data pool replay and local map update strategy in a sliding window manner.
%

For the odometry part, we extend our previous work LocNDF~\cite{wiesmann2023ral-icra} to accomplish scan-to-implicit map registration efficiently without the effort of point correspondence association using a second-order optimization to estimate the ego-motion incrementally. 
We propose an additional robust kernel based on the regularity of the predicted SDF to further improve the registration accuracy and robustness.
%

%
To correct the drift accumulated by the odometry, we use the optimized neural point local map for global loop closure detection and correction. 
Pose graph optimization~(PGO) is conducted after a loop verification to correct the drift and the inherently elastic neural points are transformed along with the poses of their associated frames to form a globally consistent map, from which a consistent SDF and mesh can be generated.






The main contribution of this paper is a novel neural SLAM system, called PIN-SLAM, based on a point-based implicit neural (PIN) map representation that supports building large-scale globally consistent maps online, as shown in \figref{fig:mot}.
To the best of our knowledge, PIN-SLAM is the first full-fledged implicit neural SLAM system including odometry, loop closure detection, and globally consistent implicit mapping.


In sum, we make four key claims, which will be backed up in the experiments:
(i) Our SLAM system achieves localization accuracy better or on par with state-of-the-art LiDAR odometry/SLAM approaches and is more accurate than recent implicit neural SLAM methods on various datasets using different range sensors. 
(ii) Our method can conduct large-scale globally consistent mapping with loop closure thanks to the elastic neural point representation.
(iii) Our map representation is more compact than the previous counterparts and can be used to reconstruct accurate and complete meshes at an arbitrary resolution.
(iv) Our correspondence-free scan-to-implicit map registration and the efficient neural point indexing by voxel hashing enable our algorithm to run at the sensor frame rate on a single NVIDIA A4000 GPU. 
%

The open-source implementation of our approach is available at: ~\url{https://github.com/PRBonn/PIN_SLAM}.

\section{Related Work}
\label{sec:related}


\subsection{LiDAR Odometry and SLAM}
SLAM using range sensors such as LiDAR sensors has been an active research topic for the last decades.
A LiDAR SLAM system often consists of odometry, mapping and optionally the loop closure correction.   
At the core of such a system are the map representation and the point cloud registration algorithm.

The earlier works on 2D LiDAR SLAM~\cite{grisetti2007tro, kohlbrecher2011ssrr, hess2016icra} adopt a probabilistic occupancy grid map and use 2D scan matching based on iterative closest point~(ICP) algorithm~\cite{besl1992pami} or particle filters~\cite{montemerlo2002aaai, grisetti2007tro} for ego-motion estimation. 

For 3D LiDAR odometry and mapping, similar to the feature-matching-based methods popularized in visual SLAM, the seminal work LOAM~\cite{zhang2014rss} proposes to extract sparse planar or edge feature points from the scan point cloud and registers them to the last frame or the feature point map using ICP.
LOAM inspired multiple follow-up works~\cite{shan2018iros, wang2021iros-fflo,pan2021icra-mvls,lin2019iros-larl,qin2020icra-lins,shan2020iros} based on more sophisticated feature point extraction~\cite{shan2018iros, wang2021iros-fflo}, classification and registration optimization schemes~\cite{pan2021icra-mvls,lin2019iros-larl} or the fusing with inertial measurements~\cite{qin2020icra-lins,shan2020iros}, leading to faster, more accurate and more robust odometry estimation. 

Recently, CT-ICP~\cite{dellenbach2022icra} and KISS-ICP~\cite{vizzo2023ral} achieved strong LiDAR odometry performance without feature point extraction. 
They register and append voxel-downsampled point cloud to a local point cloud map supporting efficient neighborhood search using either point-to-plane or point-to-point metrics. 
Combined with a motion model providing initial poses, these feature-free ``direct" methods~\cite{koide2021icra,dellenbach2022icra,di2022iros,vizzo2023ral} are comparably robust in structure-less scenarios and can take advantage of denser observations for optimization. 
The same idea has been applied to LiDAR-inertial odometry~\cite{xu2022tro, wu2024icra} at a high operating frequency.
Our method belongs to the ``direct"  algorithms but avoids the time-consuming and non-differentiable correspondence association procedure of ICP.

Except for the point-based map used by aforementioned methods, some other works estimate the ego-motion by conducting scan registration to other explicit map representations such as 3D surfels~\cite{behley2018rss, chen2019iros}, triangle meshes~\cite{vizzo2021icra,ruan2023icra}, voxelized Gaussian distributions~\cite{koide2019ijars, masashi2021icra} as well as the implicit map representations such as moving least square model~\cite{deschaud2018icra}, Gaussian process~\cite{ruan2020iros} and grid-based implicit neural SDF~\cite{deng2023iccv}.
In contrast, our method uses a point-based implicit neural SDF map representation, which combines the best of both worlds: it retains the flexibility and elasticity characteristic of point-based methods while offering continuity and compactness simultaneously thanks to the neural representation. 

There exists also learning-based LiDAR odometry methods~\cite{li2019cvpr-lonet, wang2021cvpr-pwclo, wang2022pami}, which realize registration between adjacent frames using end-to-end pose supervision. 
However, they typically cannot generalize well to unseen test sets and do not make use of the map for odometry.
Similar to recent work Nerf-LOAM~\cite{deng2023iccv}, our method exploits neural networks to fit the SDF online using the input point clouds, thus requiring no pre-training and generalizing well in various scenarios.

Besides incremental pose estimation, loop closure detection and correction are necessary for building a globally consistent, long-term SLAM system. 
Various approaches have been proposed to first conduct LiDAR place recognition and then estimate the transformation between the queried and retrieved scan using geometry-based~\cite{kim2018iros,kim2022tro} or learning-based~\cite{angelina2018cvpr, ma2022ral, luo2023iccv} scan-wise global descriptor matching as well as local feature matching and verification~\cite{luo2021ral}. 
However, previous methods often rely on the raw point cloud from a sparse single scan for descriptor generation and matching. 
In contrast, our method makes use of the neural point features in the local maps for loop closure detection. 
The usage of the local map makes our method robust to occlusions and sensors with a narrow field of view. 
Besides, reusing the online-optimized neural point features can avoid the generalizability concern of offline-trained feature extractors and save additional computation for local feature extraction.

\subsection{Implicit Neural Map Representation}

Over the past decades, explicit map representations have been widely used in the robotics community for localization~\cite{thrun2001ai}, planning, and exploration~\cite{stachniss2005rss}. 
These methods explicitly represent the scene using point cloud~\cite{zhang2014rss}, surfels~\cite{behley2018rss}, triangle meshes~\cite{vizzo2021icra}, or voxel grids storing either the occupancy probabilities~\cite{grisetti2007tro} or a truncated signed distance function~(TSDF)~\cite{newcombe2011ismar}.
Various approaches have been proposed to improve the scalability and efficiency of the map data structure~\cite{hornung2013ar, niessner2013tog,vizzo2022sensors} and to realize the incremental integration from sensor measurements~\cite{oleynikova2017iros} for the dense volumetric mapping. 
%
These explicit representations often discretize the scene with a fixed spatial resolution.
Another line of work adopts a continuous implicit representation such as Gaussian processes~\cite{wu2023tro} or reproducing kernel Hilbert maps~\cite{ramos2016ijrr}.
However, these approaches currently do not scale well to 3D data.

Recently, implicit neural representations have been proven effective in modeling radiance fields~\cite{mildenhall2020eccv} and geometric (occupancy or distance) fields~\cite{mescheder2019cvpr, park2019cvpr} using fairly simple neural networks. 
These representations have demonstrated notable success in various applications, including novel view synthesis, 3D surface reconstruction, and shape completion. 
Following the seminal works on neural radiance field (NeRF)~\cite{mildenhall2020eccv}, DeepSDF~\cite{park2019cvpr}, and occupancy networks~\cite{mescheder2019cvpr}, numerous works target improving the scalability of the implicit neural representation while boosting the time and memory efficiency. 
Instead of representing the whole scene with a single MLP, more recent methods exploit a hybrid representation by jointly optimizing explicitly stored local latent features and a shallow MLP. 
These works propose to store the optimizable local latent features in various data structures such as multi-resolution dense voxel grids~\cite{peng2020eecv}, sparse voxel hashing grids~\cite{mueller2022acmgraphics, li2023cvpr-neuralangelo}, octree nodes~\cite{takikawa2021cvpr}, permutohedral lattices~\cite{rosu2023cvpr}, axis-aligned tri-plane 2D grids~\cite{chen2022eccv-tensorf}, or as an unordered point sets~\cite{xu2022cvpr}.

The advances in efficient training of scalable implicit neural representations open up an avenue to implicit neural mapping and SLAM systems. 
Compared to traditional methods, implicit neural map representations have several attractive properties such as more compact storage, better noise smoothing capabilities and stronger inpainting and hole-filling ability for sparse or occluded observations.
In the realm of mapping and SLAM from a stream of RGB-D data, several works propose to use a single MLP~\cite{sucar2021iccv, ortiz2022rss, azinovic2022cvpr} and a hybrid representation combining grid-based local latent features and a shallow MLP~\cite{huang2021cvpr, zhu2022cvpr, yang2022ismar, wang2023cvpr-coslam, johari2023cvpr} to model the radiance or geometric field of the scene and simultaneously track the camera pose. 
These approaches have shown comparable tracking accuracy compared to the classic visual odometry methods and can generate a more complete and compact map, which can be reconstructed as a 3D mesh. 

Though with fewer works, a related trend can be seen in LiDAR mapping and localization research. IR-MCL~\cite{kuang2023ral} and LocNDF~\cite{wiesmann2023ral-icra} propose to localize the robot within an implicit neural distance map built by laser scans. 
To scale up the implicit neural representation to large-scale outdoor LiDAR data, SHINE-Mapping~\cite{zhong2023icra} stores local features in octree-based sparse voxels.
Towards a SLAM system, LONER~\cite{isaacson2023ral} integrates the incremental neural mapping into a LiDAR odometry front-end using ICP.
Recently, NeRF-LOAM~\cite{deng2023iccv} proposed an implicit neural LiDAR odometry and mapping system using an online optimizable octree-based feature grid.
%

Besides grid-based representations, the point-based implicit neural representation stores latent features in a set of neural points. 
It has been applied to efficient NeRF fitting~\cite{xu2022cvpr}, dynamic NeRF~\cite{abou2024wacv}, adaptive surface reconstruction~\cite{li2022cvpr-dccdif} and RGB-D SLAM~\cite{sandstrom2023iccv}. 
Though also using neural points, the recent work Point-SLAM~\cite{sandstrom2023iccv} targets only indoor RGB-D SLAM and does not support loop closure and globally consistent mapping. 
Point-SLAM takes advantage of the flexibility of point-based representation for the adaptive scene encoding, \ie, allocating more features to structures that require more details.
We, instead, make use of the elasticity property of neural points for building a globally consistent map online by transforming the neural points on loop closures.

The aforementioned neural implicit RGB-D and LiDAR SLAM systems still have two major limitations. 
First, most of them~\cite{sucar2021iccv, zhu2022cvpr, sandstrom2023iccv, deng2023iccv} cannot run at the sensor frame rate, limiting their applications for online robotics applications.
This is mainly due to the map data structure, computationally demanding differentiable rendering-based optimization for mapping, or an inefficient gradient descent optimization for pose estimation.
Second, the map representations of previous neural implicit SLAM approaches do not support loop closure correction, and thus cannot build globally consistent maps, especially in outdoor long-term robotics missions.
The reason for this is the usage of the regular latent feature grids that are not elastic to pose corrections introduced by loop closures.
Once a loop is corrected and the poses are updated, previous methods need to re-allocate the feature grids and re-train the entire map, which is computationally demanding and clearly prevents online applications.
To solve the first challenge, we realize an efficient SLAM system by adopting direct point-wise SDF supervision in a local map during mapping and a second-order on-manifold optimization for odometry estimation.  
To deal with the second limitation, our approach exploits the elastic and deformable point-based implicit representation to avoid grids and thus the remapping after loop correction. 

\subsection{Global Consistency in SLAM}

Global consistency is a desired property for maps. 
%
The solutions to globally consistent mapping can be divided into remapping-based, submap-based, and point-based methods.

%
The remapping-based method such as Bundle Fusion~\cite{dai2017tog} accomplishes online globally consistent 3D reconstruction using on-the-fly surface re-integration in a bundle-adjustment manner at the cost of substantial computational efforts. 

The submap-based methods accumulate the observations as a submap and assume the submap is a locally-defined rigid body. 
The pseudo-global consistency is maintained by optimizing a graph linking submaps and their associated poses through submap-to-submap registrations.
The submap can use different representations, such as feature points~\cite{pan2021icra-mvls,liu2023ral} and TSDF~\cite{reijgwart2019ral, wang2021icra}.
The usage of submaps decreases the number of nodes and edges in pose graph optimization, thus saving computational effort.
However, these methods have issues of ambiguity in the overlapping region of the adjacent submaps and determining appropriate criteria for submap division to balance the rigidity and efficiency.
%

%
The point-based methods represent the map as a fully deformable point set or surfels, each associated with a frame pose. 
Whelan~\etal~\cite{whelan2015rss} proposes Elastic Fusion for globally consistent RGB-D SLAM using surfel map with deformation graph, which is further scaled for outdoor surfel-based LiDAR SLAM~\cite{behley2018rss, chen2019iros}.
These methods can handle globally consistent mapping without the need for submap division. 
However, in contrast to volumetric mapping, point-based methods are less suitable for online 3D reconstruction and path planning. This is attributed to their sparsity and the absence of a direct representation of free or unknown space.

Only a few recent works are trying to solve the loop closure correction and globally consistent mapping using implicit neural map representation without storing the raw observation data and conducting remapping after the pose update. 
IMT-Mapping~\cite{yuan2022ral} uses the $\mathrm{SE}(3)$-equivalent grid-based feature for implicit mapping.
%
When a loop is corrected, IMT-Mapping can directly get the updated grid features by interpolating the transformed $\mathrm{SE}(3)$-equivalent features. 
Following the submap-based explicit mapping systems~\cite{reijgwart2019ral, pan2021icra-mvls,wang2021icra}, MIPS-Fusion~\cite{tang2023tog} and NF-Atlas~\cite{yu2023ral} employ an implicit submap with MLP or octree-based feature grid and adjust the submaps when a loop is corrected to keep the consistency. 

In contrast to previous methods, our point-based implicit neural map belongs to the point-based globally consistent mapping approach. 
It avoids the non-trivial submap division and overlapping disambiguation while mapping a continuous SDF that enables efficient 3D reconstruction.

\section{Our Approach to Globally Consistent\\ Implicit Neural SLAM}
\label{sec:main}


Our approach PIN-SLAM primarily addresses large-scale LiDAR SLAM for global map consistency, with applicability to other range sensors such as RGB-D cameras. 

\begin{figure*}[h]
  \centering  
  \includegraphics[width=0.9\linewidth]{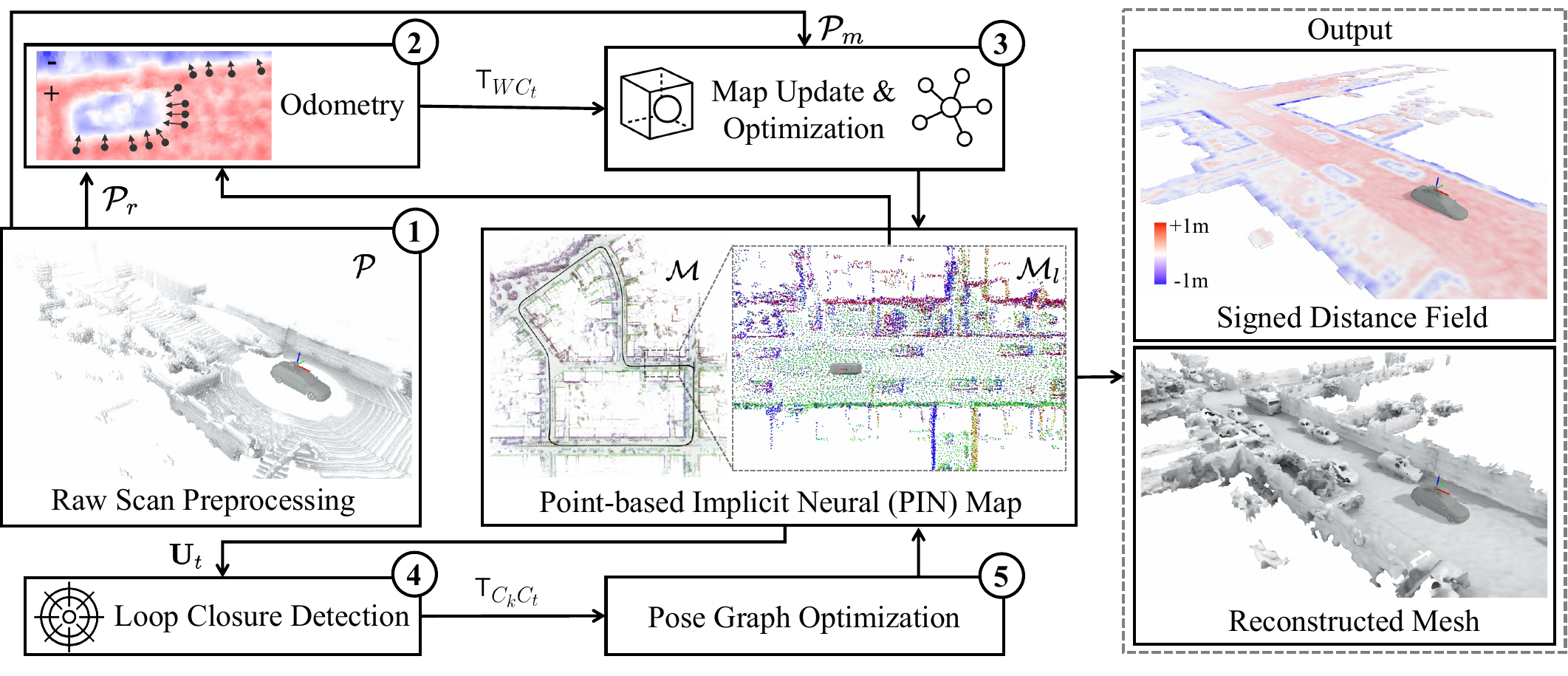}
  \setlength{\abovecaptionskip}{5pt}
  \caption{Pipeline overview of PIN-SLAM. Starting from a point cloud $\mathcal{P}$ scanned at timestep $t$, (1) a point cloud for registration $\mathcal{P}_r$ and a point cloud for mapping $\mathcal{P}_m$ are voxel-downsampled. (2) We align $\mathcal{P}_r$ to the implicit SDF of current local map $\mathcal{M}_l$ to estimate the global pose of current frame $\mq{T}_{WC_{t}}$. (3) $\mq{T}_{WC_{t}}$ is then used to transform $\mathcal{P}_m$ into the map coordinate system. With the transformed $\mathcal{P}_m$, we update the point-based implicit neural (PIN) map $\mathcal{M}$ and optimize the neural point features in the local map $\mathcal{M}_l$ by online incremental learning. (4) We generate polar context descriptor $\mathbf{U}_{t}$ using the current local map $\mathcal{M}_l$ and search for loop closures by comparing $\mathbf{U}_{t}$ to descriptors generated for previous frames. Once a loop between frame $C_{t}$ and $C_{k}$ is detected, we add the transformation $\mq{T}_{C_kC_{t}}$ as a loop edge of the pose graph and then (5) conduct the pose graph optimization. The position and orientation of the neural points in $\mathcal{M}$ are transformed along with their associated frames after the pose graph optimization, leading to a globally consistent map. With the PIN map, we can query the SDF value at an arbitrary position during or after the SLAM task for path planning and mesh reconstruction.}
  \label{fig:pipeline}
  \vspace{-8pt}
\end{figure*}

\textbf{Notation.}
In the following, we denote the transformation of a point $\vv{p}_A$ in coordinate frame $A$ to a point $\vv{p}_B$ in coordinate frame $B$ by $\mq{T}_{BA} \in \mathrm{SE}(3)$, such that $\vv{p}_B = \mq{T}_{BA} \vv{p}_A$.
Note that we use the homogeneous coordinate conversion of points before and after the affine transformation, but will not include this operation explicitly in the following derivations.
We let $\m{R}_{BA} \in \mathrm{SO}(3)$ and $\vv{t}_{BA} \in \RR^{3}$ denote the corresponding rotational and translational part of transformation $\mq{T}_{BA}$.

We denote the sensor coordinate frame at timestep $t$ as $C_t$.
Frame $C_t$ is associated to the world coordinate frame $W$ by a pose $\mq{T}_{WC_t} \in \mathrm{SE}(3)$. 
We assume the transformation $\mq{T}_{WC_0}$ from the first frame $C_0$ to the world frame $W$ is a constant, as either the identity matrix or the extrinsic calibration matrix. 
We denote the position of frame $C_t$ as $\vv{t}_{WC_t}$ and the accumulated travel distance up to frame $C_t$ as $D(t)$, which is given by:
\begin{equation}
  D(t)=\sum_{n=1}^{t} \left\|\vv{t}_{C_{n-1}C_{n}}\right\|_2.
\end{equation}


\textbf{Overview.}
Before the detailed explanation of our methodology, we provide a brief overview of PIN-SLAM's pipeline.
For each point cloud $\mathcal{P} = \{\vv{p} \in \RR^3\}$ measured at \mbox{timestep $t$}, we simultaneously estimate the pose $\mq{T}_{WC_t}$ of the sensor frame and update the point-based implicit neural map $\mathcal{M}$. 
For every timestep, as shown in \figref{fig:pipeline}, PIN-SLAM performes the following steps:

\begin{enumerate}
\item Preprocessing: We voxel-downsample the input point cloud $\mathcal{P}$ into the point cloud for registration $\mathcal{P}_r$ and the point cloud for mapping $\mathcal{P}_m$ (\secref{sec:preprocess}). 

\item Odometry: We estimate the global pose $\mq{T}_{WC_{t}}$ by registering the point cloud $\mathcal{P}_r$ to the implicit SDF of the local map $\mathcal{M}_l$. 
The odometry transformation $\mq{T}_{C_{t-1}C_{t}}$ is added as an edge in the pose graph $\mathcal{G}$ (\secref{sec:odometry}).
 
\item Mapping: We filter the dynamic points in $\mathcal{P}_m$ based on the \mbox{map $\mathcal{M}$}. 
Then we sample along the ray from the sensor to each point in $\mathcal{P}_m$ to get the training samples $\mathcal{D}$ and transform them to the world frame using $\mq{T}_{WC_{t}}$. 
We use the close-to-surface samples $\mathcal{D}_s \subset \mathcal{D}$ to initialize new neural points, append them to the map $\mathcal{M}$ and reset the local map $\mathcal{M}_l$ centered at the current position $\vv{t}_{WC_{t}}$. 
We update the training sample pool $\mathcal{D}_p$ with the training samples $\mathcal{D}$ of the current frame.
Then we optimize the neural point features in the local map $\mathcal{M}_l$ using the samples in the pool $\mathcal{D}_p$ with direct SDF supervision by gradient descent. After that, we allocate the updated local map $\mathcal{M}_l$ back to the global map $\mathcal{M}$ (\secref{sec:mapping}).

\item Loop closure detection: We generate a local polar context descriptor $\mathbf{U}_{t}$ using the local map $\mathcal{M}_l$. 
Then we search for a potential loop closure by comparing the feature distances between the descriptors of the current frame and candidate frames. 
Once a loop closure candidate between frame $C_t$ and $C_k$ is detected, we verify it by registering the point cloud $\mathcal{P}_r$ of the current frame to the local map $\mathcal{M}_l$ centered at the sensor position $\vv{t}_{WC_{k}}$ of frame $C_k$. 
We add the loop transformation $\mq{T}_{C_{k}C_{t}}$ resulting from the registration as an edge in the pose graph $\mathcal{G}$ if the registration succeeds (\secref{sec:loop}). 

\item Pose graph optimization: Once a loop closure edge is added, we optimize the pose graph $\mathcal{G}$.
The position and orientation of each neural point in the global map $\mathcal{M}$ are transformed along with its associated frames after the optimization to keep the global consistency. 
We then transform the training sample pool $\mathcal{D}_p$ and reset the local map $\mathcal{M}_l$ accordingly after loop correction (\secref{sec:pgo}). 
\end{enumerate}

For the first timestep, we initialize the map using only the first scan as done in the mapping step. During or after the SLAM, we can query the SDF value at an arbitrary position for mesh reconstruction via the marching cubes algorithm~\cite{lorensen1987siggraph}. 

Next, we will explain the basics, data structure, and training process of the proposed PIN map representation (\secref{sec:pin_map}) and then explain each step of the pipeline in more detail.

\subsection{Neural Point-based Map Representation}
\label{sec:pin_map}

\subsubsection{Map Representation}
We define the proposed point-based implicit neural map as a set of neural points: 
\begin{equation}
  \mathcal{M}= \{\vv{m}_i = \left(\vv{x}_i, \vv{q}_i, \vv{f}_i^g, t^c_i, t^u_i, \mu_i\right) \mid i=1, \ldots, N\}, 
\end{equation}
where each neural point $\vv{m}_i$ is defined in the world frame $W$ by a position $\vv{x}_i \in \mathbb{R}^3$ and a quaternion $\vv{q}_i \in \mathbb{R}^4$ representing the orientation of its own coordinate frame. 
Each neural point stores the optimizable latent feature vectors $\vv{f}_i^g \in \mathbb{R}^{F_g}$ representing the local geometry.
In addition, we keep track of the creation timestep $t^c_i$, the last update timestep $t^u_i$ and the stability $\mu_i$ for each neural point to determine whether a neural point is active or inactive, stable or unstable.
We link each neural point $\vv{m}_i$ to the sensor pose $\mq{T}_{WC_{t^m_i}}$ at the mean of the neural point's creation and last timestep $t^m_i = \lfloor (t^c_i+t^u_i)/2 \rfloor$ to directly manipulate the map by updating the sensor poses.

\begin{figure}[tp]
  \centering  
  \includegraphics[width=0.98\linewidth]{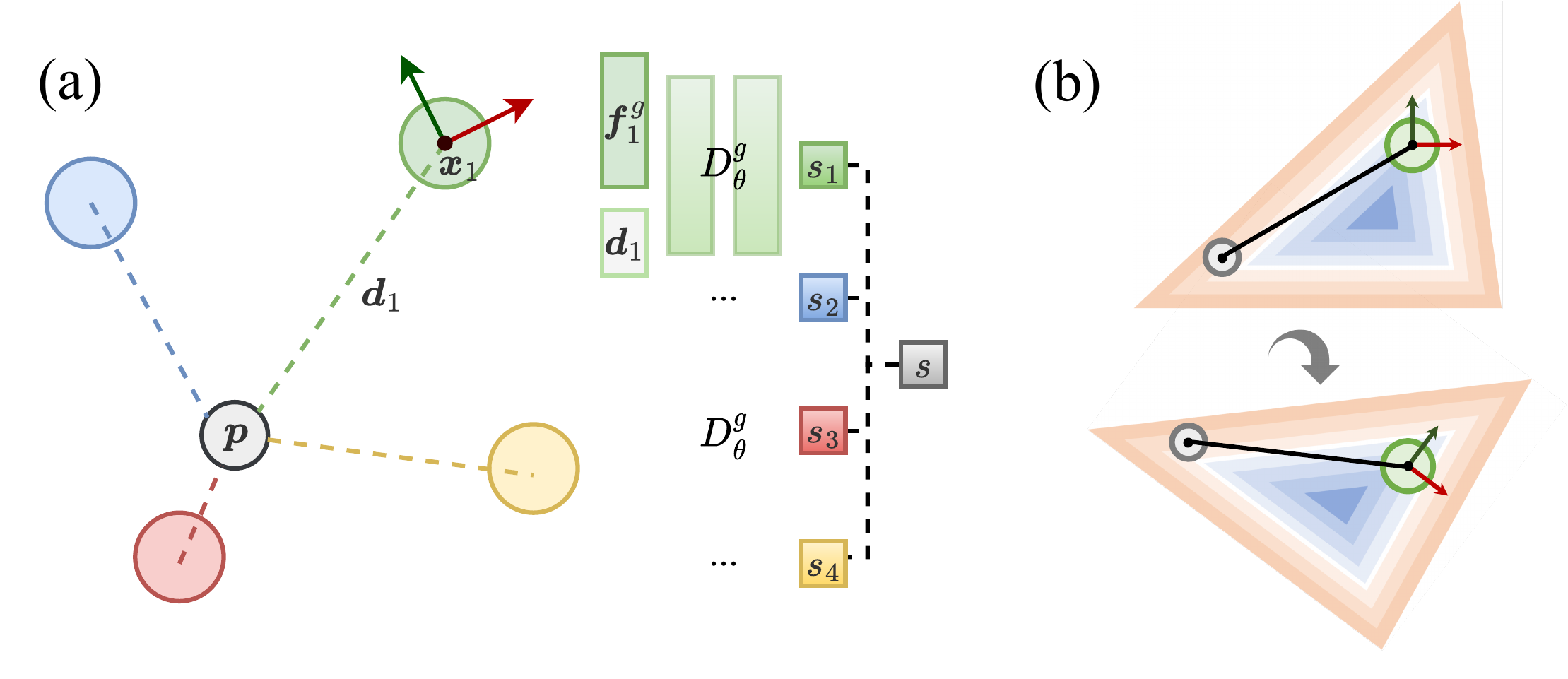}
  \setlength{\abovecaptionskip}{2pt}
  \caption{Diagram of SDF querying in our point-based implicit neural map simplified in 2D. (a) The point in gray is the query position $\vv{p}$ while the other points are the neighboring neural points. Each neural point predicts the SDF value $s_i$ at the query position by feeding the neural point feature $\vv{f}^g_i$ and the query point's position $\vv{d}_i$ under the neural point's coordinate system through a globally shared decoder $D_\theta^g$. Then the predictions are weighted as the final prediction $s$ according to the distances from the neural points to the query position. (b) The orientation of each neural point defines its local coordinate system, ensuring the relative coordinate $\vv{d}_i$ and thus the SDF querying invariant to rigid-body transformation.}
  \label{fig:mlp}
  \vspace{-8pt}
\end{figure}

As shown in \figref{fig:mlp}, similar to the auto-decoder architecture of DeepSDF~\cite{park2019cvpr}, we predict the SDF value $s$ at a query position $\vv{p}$ conditioned on $K$ nearby neural points.
For each neural point $\vv{m}_j$ in the $K$-neighborhood $\mathcal{N}_p$ of the query position $\vv{p}$, we first concatenate the latent feature vector $\vv{f}_j^g$ and the relative coordinate $ \vv{d}_j$. Here, $\vv{d}_j$ represents the query position in the coordinate system of the neural point $\vv{m}_j$.
Then we feed the concatenated vector to a globally shared geometric decoder $D_\theta^g$, a shallow MLP containing $M_{\text{mlp}}$ hidden layers with $N_{\text{mlp}}$ neurons, to predict the SDF value $s_j$ as:
\begin{equation}
s_j = D_\theta^g(\vv{f}_j^g, \vv{d}_j),
\label{equ:sdf_prediction}
\end{equation}
where the relative coordinate $\vv{d}_j$ is given by:
\begin{equation}
\vv{d}_j = \vv{q}_j (\vv{p} - \vv{x}_j)  \vv{q}_{j}^{-1}. 
\label{equ:d}
\end{equation}
As shown in \figref{fig:mlp}(b), the concatenation of relative coordinate $\vv{d}_j$ makes the prediction invariant to the local translation and rotation, thus enabling the elasticity of the map after loop closure correction, which will be further explained in~\secref{sec:pgo}.

The predicted SDF values $s_j$ of the $K$ nearest neural points at the query position $\vv{p}$ are then interpolated as the final prediction $s = S(\vv{p})$ by inverse distance weighting, given by:
\begin{equation}
  S(\vv{p}) =\sum_{j \in \mathcal{N}_p} \frac{w_j}{\sum_{k \in \mathcal{N}_p} w_k} s_j, 
\label{equ:interpolation}
\end{equation}
where the weights $w_j$ are defined as:
\begin{equation}
  w_j={\left\|\vv{p} - \vv{x}_j\right\|^{-2}}.
\label{equ:weighting}
\end{equation}

Likewise, we define the stability $\mu = H(\vv{p})$ at the query position $\vv{p}$ as a distance-weighted mean of the stability $\mu_j$ of the $K$ nearby neural points by switching $s_j$ with $\mu_j$ in \eqref{equ:interpolation}.

Intuitively, the final prediction can be regarded as a voting of the individual prediction from each neighboring neural point $\vv{m}_j$ or an ensemble of multiple locally defined DeepSDF-like auto-decoder models. 
In contrast with the relatively deep decoder used in DeepSDF~\cite{park2019cvpr}, which needed to be pre-trained to represent the object globally, we represent the local geometries mainly in the locally defined latent feature space of neural points and use a shallow decoder as a general interpreter to map from the feature space to SDF values.

\subsubsection{Map Data Structure}
For fast neural point indexing and neighborhood search, we maintain a voxel hashing data structure with a fixed voxel resolution $v_p$ and a hash table size $T$.
This structure serves to organize the neural points, ensuring that each voxel contains no more than one active neural point.
In line with prior work~\cite{mueller2022acmgraphics,vizzo2023ral}, we use a spatial hashing function $\kappa = h(\vv{p}_W)$ to map from a position $\vv{p}_W \in \RR^3$ to an entry $\kappa \in \ZZ$ in the hash table.
This entry stores either the index $i$ of a neural point $\vv{m}_i$ in the map $\mathcal{M}$ or the default value $-1$, indicating that the entry is not yet occupied.

In contrast to Point-SLAM~\cite{sandstrom2023iccv}, for efficient proximity search during the SDF prediction at a query position $\vv{p}$, we use the voxel structure to find the neural points in the voxel-defined neighborhood $\mathcal{N}_{p}^{v}$ containing $N_n \times N_n \times N_n$ voxels centered at the corresponding voxel of $\vv{p}$ with constant time access.
We sort the distances and take the $K$ nearest neural points in $\mathcal{N}_{p}^{v}$ to get the $K$-neighborhood $\mathcal{N}_{p}$ for SDF prediction.
The larger $N_n$ is, the larger the receptive field would be while the computation time for neighborhood search would increase. 
If the number of neighboring neural points $K_n$ is $1\leq K_{n}<K$, we conduct SDF interpolation among the $K_n$ points.

\subsubsection{Map Initialization and Update}
We initialize or update neural points at each timestep using the measured point cloud.  
For a point $\vv{p}_W$ in the world frame $W$ measured at timestep $t$, we will initialize a new neural point $\vv{m}_k$, append it to the map $\mathcal{M}$ and set the hashing entry's value as the new neural point's index, \ie, $\kappa \leftarrow k$ only under the following three circumstances: 
\begin{enumerate}
  \item The hashing entry is not yet occupied, \ie, $\kappa=-1$. 
  \item There is a hash collision, \ie, the position $\vv{x}_{\kappa}$ of currently stored neural point $\vv{m}_{\kappa}$ in the voxel of the added point $\vv{p}_W$ is far away from $\vv{p}_W$ due to hash collision. 
  \item The stored neural point $\vv{m}_{\kappa}$ is no longer active, \ie, $D(t)-D(t^u_{\kappa})>d_l$, meaning the travel distance from the last updated timestep $t^u_{\kappa}$ of the neural point to the current timestep $t$ is larger than a travel distance threshold $d_l$. This is used to differentiate between the historical observation and the present observation at the same position in the world frame upon revisiting. 
\end{enumerate}

In such cases, we initialize a new neural point $\vv{m}_k$ to have a position $\vv{x} = \vv{p}_W$, an identity quaternion $\vv{q}=\left(1,0,0,0\right)$, a zero feature vector $\vv{f}^{g}=\vv{0}$, creation and last update timestep as $t^{c}=t^{u}=t$, and the stability as $\mu=0$.
In the latter two cases, the originally stored neural point is replaced by the new one and no longer indexed by the voxel hashing map. 
However, they are kept in the global neural point map for the sake of loop closure correction (\cf \secref{sec:loop}) and globally consistent map adjustment (\cf \secref{sec:pgo}).

\subsubsection{Local Map}
%
%
Our method distinguishes between an active local map $\mathcal{M}_l$ and a global map $\mathcal{M}$. The registration described in \secref{sec:odometry} and \secref{sec:loop} are performed using the local map $\mathcal{M}_l$ to avoid the alignment to the inconsistent historical observations caused by odometry drift.

In previous works, a local map is often defined using either a spatial window~\cite{pan2021icra-mvls,vizzo2023ral} or a temporal window~\cite{whelan2015rss,behley2018rss}.
We propose to use both of them.
In practice, the speed of the robot may vary a lot during the operation. 
For instance, the robot may stop at a position for a long time.
Since the drift of the odometry is often proportional to the accumulated travel distance $D(t)$ instead of the time span, we propose to use a travel distance threshold instead of the timestep threshold~\cite{whelan2015rss,behley2018rss} to determine the local temporal window.
Consequently, we define a local map $\mathcal{M}_l$ centered at the sensor position $\vv{t}_{WC_t}$ at timestep $t$. 
This map includes all neural points within both the spatial window $\left\|\vv{x_i} - \vv{t}_{WC_t}\right\|_2<r_l$ and the temporal window $D(t) - D(t_i^c) < d_l$. 
Here, $r_l$ represents the radius of the local spatial window, and $d_l$ denotes the previously mentioned travel distance threshold.

Next, we will explain how we incrementally optimize the neural point features in the local map $\mathcal{M}_l$ using a sliding window-like training sample pool $\mathcal{D}_p$.


\subsubsection{Map Training Samples}
%
%
At each timestep $t$, we take samples along the rays from the voxel-downsampled egocentric scan $\mathcal{P}_m = \{\vv{p} \in \RR^3\}$ to collect training data. 
On each ray $\vv{r} =\vv{p}/\left\|\vv{p}\right\|_2$ from the sensor to the measured point $\vv{p}$, we can represent a sampled point $\vv{u}$ in the sensor frame $C_t$ uniquely with its depth $d$ along the ray, given by: $\vv{u} = d \vv{r}$.
We take the endpoint ($d=\left\|\vv{p}\right\|_2$) and $N_s$ points close to the surface with the depth sampled from a Gaussian distribution $d_s \sim \mathcal{N}\left(\left\|\vv{p}\right\|_2, \sigma_s^2\right)$ with the endpoint as the mean and $\sigma_s$ as the standard deviation. 
Additionally, we sample $N_f$ points uniformly in the free space in front of the surface with the depth $d_{f} \sim U\left(\zeta_{\text{min}}\left\|\vv{p}\right\|_2,\left\|\vv{p}\right\|_2-2\sigma_s\right)$ and $N_b$ points in the truncated free space behind the surface with the depth $d_{b} \sim U\left(\left\|\vv{p}\right\|_2+2\sigma_s,\left\|\vv{p}\right\|_2+d_{b}\right)$, where $\zeta_{\text{min}}$ is the minimum sample depth ratio and $d_{b}$ represents the maximum sample range behind surfaces.
For each sample point $\vv{u}$, we take the projective signed distance along the ray as its SDF target value $\hat{s} = \left\|\vv{p}\right\|_2 - d$.
Though the projective distance would always be an overestimation, it is computed quickly and is a good approximation when the sample point is close to the surface. 

We define the training samples $\mathcal{D}$ at timestep $t$ as $N_t$ sample positions and their target values from all the rays, given by:
\begin{equation}
  \mathcal{D}= \left\{\left(\vv{u}_j, \hat{s_j}, t \right) \mid j=1, \ldots, N_t\right\}, 
\end{equation}
where $N_t = M_t(N_s+N_f+N_b+1)$ and $M_t$ is the number of points (rays) in point cloud $\mathcal{P}_m$.
We associate the training samples with the corresponding sensor frame $C_t$ by recording the timestep $t$ of each sample so that we can later transform each sample point into the world frame.

\begin{figure}[!t]
  \centering  
  \includegraphics[width=0.985\linewidth]{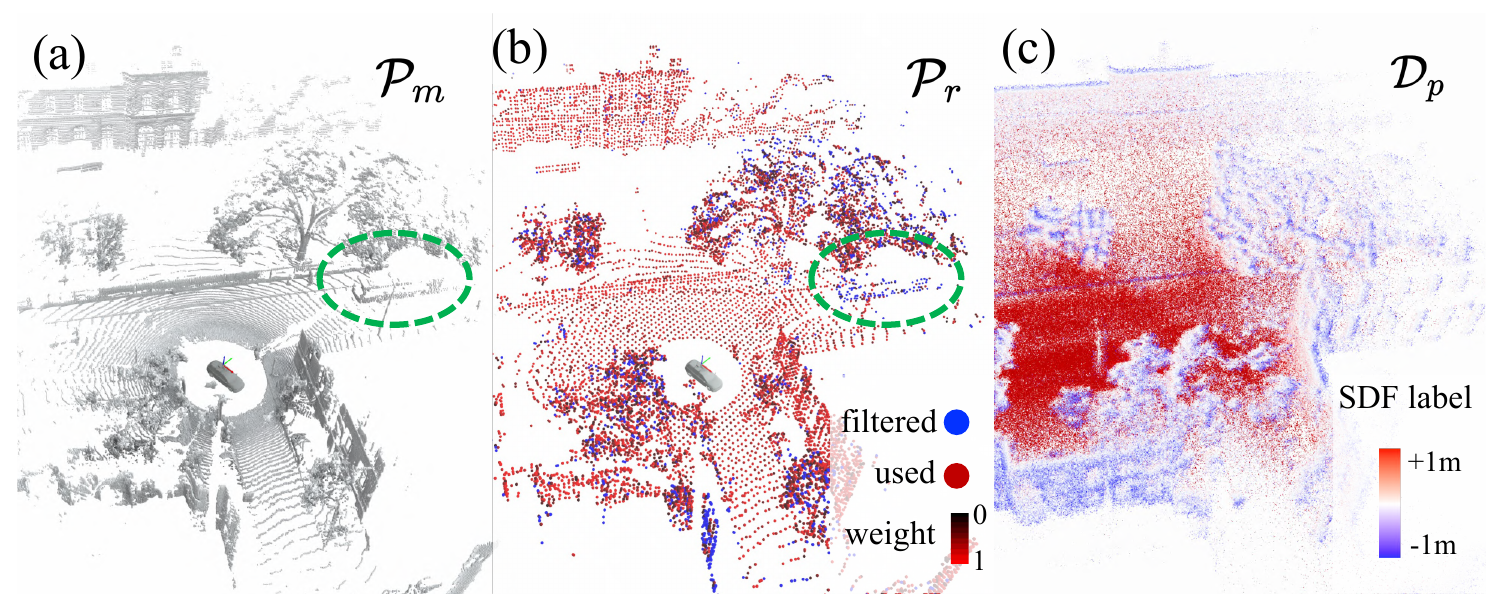}
  \setlength{\abovecaptionskip}{4pt}
  \caption{An example during the operation of PIN-SLAM at a timestep: (a) shows the point cloud for mapping $\mathcal{P}_m$. A moving bus is highlighted in the green circle. (b) shows the sparser point cloud for registration $\mathcal{P}_r$, colorized according to the point-wise registration weight from black to red. The points in blue are filtered, mainly lying at dynamic objects, rough vegetation, and newly observed regions. (c) shows points from the training sample pool $\mathcal{D}_p$ for map optimization, colorized according to their SDF target values from blue to red.}
  \label{fig:sample_scan}
  \vspace{-8pt}
\end{figure}

To tackle the ``catastrophic forgetting" problem in incremental mapping, we maintain a training sample pool $\mathcal{D}_p$ by appending the samples $\mathcal{D}$ from each timestep for replaying the historical samples, as shown in \figref{fig:sample_scan}(c). 
We filter $\mathcal{D}_p$ at each timestep $t$ by removing the samples lying outside a local sliding window with a radius $r_p$, centered at the current sensor position $\vv{t}_{WC_t}$, given by: 
\begin{equation}
  \left\|\vv{u_W^i} - \vv{t}_{WC_t}\right\|_2>r_p,
\end{equation}
where $\vv{u_W^i} = \mq{T}_{WC_{t_i}}\vv{u_i}$ is the sample position in world frame and $r_p = r_l-\frac{\sqrt{3}}{2}
N_{n}v_p$ is set to enforce that the training samples would only affect the neural points in the local map, considering the local map radius $r_l$, neighborhood voxel counts $N_n$, and the voxel size $v_p$. 
Additionally, to deal with the case when the robot stops, if the number of samples in $\mathcal{D}_p$ exceeds $N_p$, we randomly keep $N_p$ samples.

\subsubsection{Map Training Losses}
%
%
We want to train our neural points to predict SDF value on arbitrary locations and these neural point features need to be trained ideally during the incremental mapping process.
During incremental mapping, we randomly sample from the training sample pool $\mathcal{D}_p$ in batch for training. 
For the training of the SDF, we use a similar loss function as our previous work~\cite{zhong2023icra}, which combines the binary cross entropy~(BCE) loss and the Ekional regularization loss.

At a sample position $\vv{u_W^i} = \mq{T}_{WC_{t_i}}\vv{u_i}$ in the world frame, we map both the SDF prediction $s_i=S(\vv{u_W^i})$ from the nearby neural points and target value $\hat{s}_i$ to the range of $\left[0, 1\right]$ by a scaled sigmoid function $\Phi_s(s) =1 /(1+e^{s / \sigma_{t}})$ as:
\begin{equation}
  o_i=\Phi_s(s_i), ~~~ \hat{o}_i = \Phi_s(\hat{s}_i),
\end{equation}
and then the BCE loss is calculated for $N$ samples as:
\begin{equation}
  \label{equ:bce}
  \mathcal{L}_\text{bce} = \frac{1}{N} \sum_{i=1}^N \hat{o}_i \log(o_i) +  (1-\hat{o}_i) \log(1 - o_i).
\end{equation}
The BCE loss enables fast convergence of the SDF~\cite{zhong2023icra}. 
It accomplishes the logistic regression of the projective SDF value and realizes the soft truncation of signed distance close to the surface. 
We can adjust the truncation smoothness using the scale factor $\sigma_{t}$ in the scaled sigmoid function $\Phi_s$.

One necessary property of SDF is the Ekional equation~\cite{gropp2020icml}, \ie, $\left\|\nabla S\left(\vv{x}\right)\right\|_2=1$. 
Therefore, we use the Ekional loss $\mathcal{L}_{\text{eik}}$ to enforce the regularity and validity of the fitted SDF:  
\begin{equation}
  \label{equ:Ekional}
  \mathcal{L}_{\text{eik}}=\frac{1}{N} \sum_{i=1}^N \left(\left\|\nabla S\left(\vv{u}_W^i\right)\right\|_2-1 \right)^2.
\end{equation}
%
%
In line with Neuralangelo~\cite{li2023cvpr-neuralangelo}, we calculate the numerical gradient by manually adding perturbations instead of using the analytical gradient based on automatic differentiation. 
This enables the backpropagation updates beyond the local hash grid that stores neural points, resulting in a smoother SDF gradient for the calculation of $\mathcal{L}_{\text{eik}}$. 
The gradient component at a position $\vv{x}$ on the $x$ axis is given by:
\begin{equation}
  \label{equ:gradient}
  \nabla_x S(\vv{x}) = \frac{S(\vv{x}+\vv{\epsilon_x})-S(\vv{x}-\vv{\epsilon_x})}{2\epsilon},
\end{equation}
where $\boldsymbol{\epsilon_x} = (\epsilon,0,0)^\top$ is the perturbation vector on the $x$ axis and $\epsilon$ is the perturbation step size used on every axis. 
The gradient on the $y$ and $z$ axis can be calculated likewise.

The final loss is then formulated as a weighted sum of BCE and Ekional term, given by:
\begin{equation}
  \label{equ:trainloss}
  \mathcal{L} =  \mathcal{L}_\text{bce} + \lambda_{e} \mathcal{L}_\text{eik},   
\end{equation} 
where $\lambda_{e}$ is the weight for Ekional loss.
During training, we minimize the loss by optimizing latent features of the involved neural points in the local map. 
For each sample point $\vv{u}_i$, we also update the last update timestep $t^u_j$ and stability $\mu_j$ of each neural point $\vv{m_j}$ in the neighborhood $\mathcal{N}_p$ of $\vv{u}_i$, given by: 
\begin{equation}
  t^u_j \leftarrow \max(t^u_j, t_i), ~~~ \mu_j \leftarrow \mu_j + \frac{w_j}{\sum_{j \in \mathcal{N}_p} w_j}, 
\end{equation}
where $t_i$ is the timestep of this training sample, and the \mbox{weight $w_j$} can be calculated as \eqref{equ:weighting}.

Next, we will explain how we use the proposed map representation, namely the PIN map, to realize a LiDAR SLAM system towards globally consistent mapping.

\subsection{Preprocessing}
\label{sec:preprocess}

For each egocentric point cloud $\mathcal{P}$ measured at timestep $t$, we voxel-downsample $\mathcal{P}$ into the point cloud for mapping $\mathcal{P}_m$ with a smaller voxel size $v_m$, and the point cloud for registration $\mathcal{P}_r$ with a larger voxel size $v_r$.  
During the downsampling, we keep exactly one point, namely the one whose coordinate is the closest to the voxel center.

Typically for LiDAR sensors, the point cloud of one sweep~(frame) consists of the continuously scanned points measured across the acquisition time.
Therefore, for each scan frame, we first need to revert the distortions of the point cloud caused by the sensor motion.
In line with Vizzo~\etal~\cite{vizzo2023ral}, we use a constant velocity model for motion prediction.
Based on this model, we first deskew the points  $\mathcal{P}_r$ for registration before the odometry estimation by interpolating the motion prediction with the pointwise timestamp.
We deskew $\mathcal{P}_m$ with the more accurate ego-motion from odometry afterward.
%
%

\subsection{Odometry Estimation}
\label{sec:odometry}

To achieve efficient and robust odometry, we propose a correspondence-free, scan-to-implicit map registration method based on second-order optimization under multiple weighting strategies, which is based on the approach proposed by Wiesmann \etal~\cite{wiesmann2023ral-icra} targeting localization in known maps.

Our goal is to align the source point cloud $\mathcal{P}_r$ at \mbox{timestep $t$} to the neural SDF of the local PIN map $\mathcal{M}_l$ by finding the transformation $\mq{T}^* \in \mathrm{SE}(3)$ minimizing the least square error of the SDF prediction at the transformed points, given by:
\begin{equation}
  \mq{T}^*=\underset{\mq{T}}{\operatorname{argmin}} \sum_{\vv{p} \in \mathcal{P}_r} S\left(\mq{T}\vv{p}\right)^2,
\end{equation}
which can be solved by a Levenberg-Marquardt optimization taking the initial guess predicted by a constant velocity model.

We denote the $6$DOF transformation parameters as the Lie algerba $\vv{\xi}=\left[\vv{t}, \vv{\Theta}\right] = \log(\mq{T})$, where $\vv{t}$ is the translation vector and $\vv{\Theta} = \log(\m{R})$ is the axis-angle representation of the rotation matrix $\m{R}$. 
The Jacobian of the transformed point $\vv{p}_i^\prime = \mq{T}\vv{p}_i$ with regards to $\vv{\xi}$ is given by: 
\begin{equation}
  \m{J}_i=\left[\frac{\partial S\left(\vv{p}_i^\prime\right)}{\partial \vv{t}}, \frac{\partial S\left(\vv{p}_i^\prime\right)}{\partial \vv{\Theta}}\right]=\left[\vv{g}_i^{\tr}, (\vv{p}_i^\prime \times \vv{g}_i)^{\tr}\right],
  \label{equ:jacobian}
\end{equation}
where $\vv{g}_i = \nabla S\left(\vv{p}_i^\prime\right)$ is the distance gradient at $\vv{p}_i^\prime$, which can be queried from the implicit neural distance field of PIN map by automatic differentiation.
In our experiments, we discovered that using this analytical gradient for odometry estimation enhances robustness compared to utilizing the smoother but less precise numerical gradient employed for the Ekional regularization in mapping, see \eqref{equ:gradient}.

Intuitively, the registration can be solved by knowing in which direction, given by $\vv{g}$ and how much, given by $S(\vv{p^\prime})$, we have to go. This is conducted without knowing the explicit point-to-point correspondences.
The \textit{de facto} optimization target in previous neural implicit SLAM approaches~\cite{sucar2021iccv,zhu2022cvpr,yang2022ismar,sandstrom2023iccv} is the depth rendering loss, which can be seen as related to the point-to-point metric with projection-based data association. 
Our method optimizes the point-to-model SDF loss, which is similar to the point-to-plane metric in ICP with the closest surface-based data association.
The latter method has a faster convergence rate and more robust optimization according to the study by Rusinkiewicz, and Levoy~\cite{rusinkiewicz2001dim}.

We then approximate the Hessian matrix $\m{H}$ as $\m{H}=\m{J}^{\tr} \m{P} \m{J}$ and calculate the gradient of the target function as $\vv{g}=\m{J}^{\tr} \m{P} \vv{b}$, where $\m{J} \in \RR^{N_r \times 6}$ is the Jacobian, $\m{P}\in \RR^{N_r \times N_r}$ is the weight matrix, and $\vv{b}\in \RR^{N_r}$ is the residual vector.
We filter points that have fewer than $K$ neural points in their $N_p^v$ neighborhood and set each row of $\m{J}$ using the $N_r$ remaining points as \eqref{equ:jacobian}.  

For each iteration, the increment of the transformation parameters $\delta \vv{\xi}$ is given by:
\begin{equation} 
  \delta \vv{\xi} = \big(\m{H} + \lambda_d \mathrm{diag}\left(\m{H}\right)\big)^{-1} \vv{g},
\label{equ:lm}
\end{equation}
where $\lambda_d$ is the damping parameter for Levenberg-Marquardt optimization. 
The termination criterion is determined by a threshold $\gamma_c$ for the applied correction $\delta \vv{\xi}$ and a maximum number of iterations $\tau_c$.

To robustify the optimization under noisy measurements and underfitted map, we apply the Geman-McClure (GM) robust kernel for both the SDF residual $S(\vv{p}_i^\prime)$ and the SDF gradient anomaly $\epsilon_i$ to down-weight the potential measurement noise, dynamic objects, and underfitted details of the implicit map. The pointwise robust kernel weights are given by:
\begin{align}
  w_i^r=&\left(\frac{\kappa_r}{\kappa_r^2+S(\vv{p}_i^\prime)^2}\right)^2, \\
  w_i^g=&\left(\frac{\kappa_g}{\kappa_g^2+\varepsilon_i^2}\right)^2,
\end{align}
%
where $\kappa_r$ and $\kappa_g$ are the scale parameters of the robust kernel. The gradient anomaly $\varepsilon_i$ is defined as the deviation of the distance gradient from the Ekional equation, given by:
\begin{equation}
\varepsilon_i = \big|\left\|\nabla S\left(\vv{p}_i^\prime\right)\right\|_2-1\big|.
\end{equation} 

To apply the robust kernel to the optimization, we scale each diagonal element of the weight matrix $\m{P}$ with the point-wise robust kernel weights $w_i^r w_i^g$.
An example of the filtering and point-wise weight of $\mathcal{P}_r$ is shown in \figref{fig:sample_scan}(b).

%
After convergence, we additionally calculate the Eigenvalues~$\vv{\lambda}$ of the Hessian matrix $\m{H}$ for a degeneracy check~\cite{zhang2016icra}. 
The registration will be regarded as a success if the average residual $\bar{\vv{b}}>b_s$, the ratio of valid points $\alpha > \alpha_s$, and the minimum Eigenvalue  $\min(\vv{\lambda}) > \lambda_s$ for the last iteration, where $b_s$, $\alpha_s$, and $\lambda_s$ are the thresholds.
If the optimization succeeds, \ie, all checks are fulfilled, we take $\mq{T}_{WC_t}=\exp(\vv{\xi})$ as the estimated pose of the current frame $C_t$ for mapping (\secref{sec:mapping}). Otherwise, we directly use the initial guess, \ie, the constant velocity estimate, but skip the mapping step to avoid introducing wrong measurements. 

For a potential pose graph optimization, we add the odometry transformation $\mq{T}_{C_{t-1}C_{t}}$ as an odometry edge connecting pose node $C_{t-1}$ and $C_{t}$ in the pose graph $\mathcal{G}$.

\subsection{Mapping and Local Bundle Adjustment}
\label{sec:mapping}

With the sensor poses $\mq{T}_{WC_t}$ estimated by the odometry at each timestep, we can update and optimize the PIN map $\mathcal{M}$ using the point cloud for mapping $\mathcal{P}_m$.

Firstly, we filter the dynamic objects from $\mathcal{P}_m$ and use only the static part for mapping.  
Inspired by Dynablox~\cite{schmid2023ral-dynablox}, we make use of the assumption that a measured point lying in the stable free space can be regarded as a dynamic point. 
Thereby, we filter points fulfilling $\big(S\left(\vv{p}_W\right)>\gamma_d\big) \wedge \big(H\left(\vv{p}_W\right)>\gamma_{\mu}\big)$, where $S\left(\vv{p}_W\right)$ and $H\left(\vv{p}_W\right)$ are the SDF and stability prediction at the measured point $\vv{p}_W = \mq{T}_{WC_t}\vv{p}$ in the world frame. 
$\gamma_d$ and $\gamma_{\mu}$ are the free space distance and stability threshold, respectively. 
Only the static points remain for mapping.

As we explained in \secref{sec:pin_map}, we get training samples $\mathcal{D}$ by sampling along the rays from the sensor to the static points in $\mathcal{P}_m$.
We initialize new neural points in the map $\mathcal{M}$ using the close-to-surface sample points $\mathcal{D}_s$ transformed by current \mbox{pose $\mq{T}_{WC_{t}}$}. 
We then reset the local map $\mathcal{M}_l$ centered at the current position $\vv{t}_{WC_{t}}$, and append current training samples $\mathcal{D}$ to the training sample pool $\mathcal{D}_p$.
Next, we conduct PIN map optimization by gradient descent using the training samples from $\mathcal{D}_p$ in batches. 
When the SLAM starts, we jointly optimize the neural point feature $\vv{f}^g$ in the local map $\mathcal{M}_l$ and the weights of the MLP decoder $D_\theta^g$ for the first $F_{\text{mlp}}$ timesteps.
After $F_{\text{mlp}}$ timesteps, we freeze $D_\theta^g$ and optimize only the neural point features to avoid catastrophic forgetting due to the ever-changing decoder while incremental mapping.

The odometry estimation relies on pairwise scan-to-map registration and always optimizes only the latest pose.
It ignores the multi-view consistency of multiple scans in the local sliding window, thus pruning to rapidly accumulated drift. 
Inspired by the bundle adjustment (BA) technique widely employed in photogrammetry and computer vision, Liu \etal~\cite{liu2023tro} introduce a LiDAR bundle adjustment algorithm. 
This algorithm aims to refine odometry by simultaneously optimizing the scene geometry and a collection of sensor poses. 
The target of this optimization is to minimize inconsistencies among explicit geometric features such as edges and planes. 

We propose an implicit local bundle adjustment approach making use of the implicit neural map without the tedious extraction of explicit geometric features. 
For every $F_{\text{ba}}$ timesteps, we jointly optimize the sensor poses at the past $N_{\text{ba}}$ timesteps, and the neural point features in the local map $\mathcal{M}_l$, taking the current state as the initial guess. 
We optimize the L2 SDF regression loss by gradient descent using only the endpoint samples from the training sample pool $\mathcal{D}_p$ to achieve a more consistent local map and poses. The loss function is given by:
\begin{equation}
  L_\text{ba} = \sum_{\tau \in \mathcal{T}_\text{ba}} \sum_{\vv{p}_{i} \in \mathcal{P}_m^{\tau}} S\left(\mq{T}_{WC_{\tau}}\vv{p}_i\right)^2,
\end{equation}
where $\mathcal{T}_\text{ba} = \{t-N_{\text{ba}}+1, \ldots, t-1, t\}$ are the timesteps used for local bundle adjustment. $t$ is the current timestep, and $\mathcal{P}_m^{\tau}$ is the point cloud for mapping at timestep $\tau$. 

After the map optimization and local bundle adjustment, we assign the updated local map $\mathcal{M}_l$ back to the global map $\mathcal{M}$.

\subsection{Loop Closure Detection}
\label{sec:loop}

Detecting loop closures is essential to correct the accumulated drift of the odometry for globally consistent mapping.
%
%

Firstly, we use a distance-based criterion for identifying local loop closures. 
Specifically, we assess whether the positions $\vv{t}_t$ and $\vv{t}_{t_i}$ of the current scan at $C_t$ and a historical scan at $C_{t_i}$ (where $t_i < t$ and $D(t) - D(t_i) > d_l$) meet the condition: 
\begin{equation}
  \left\|\vv{t}_t - \vv{t}_{t_i}\right\|_2 < d_\text{loop}.
\end{equation}
If there is no local loop candidate, we search for global loop closures by comparing frame-wise descriptors. 
We propose a neural point feature enhanced local map descriptor for typical driving scenarios. 
This descriptor is used for global loop detection and provides a semi-metric localization as the initial guess for the relative loop transformation.
To address the place recognition challenges arising from diverse scan patterns and occlusions caused by varied viewpoints and dynamic objects, we use the local map as the processing unit for loop detection, instead of relying on a single scan.
Therefore, at each timestep $t$, with a latency of $F_{\text{lat}}$ timesteps waiting for sufficient map training using the most recent observations, we generate a local context descriptor $\mathbf{U}_{t}$ using neural points in the local map $\mathcal{M}_l$.

Similar to Scan Context~\cite{kim2018iros}, the neural points in the local map are mapped into discretized 2D polar-coordinate bins by a bird-eye-view projection with the ring resolution $H_r$ and sector resolution $H_s$.
Instead of encoding the maximum point height in each bin to create the descriptor~\cite{kim2018iros}, we encode the average neural point features optimized by the self-supervised online mapping and keep a $F_g$ dimensional vector in each bin, thus resulting in a descriptor $\mathbf{U}_{t} \in \RR^{H_r \times H_s \times F_g}$.
We average the sector dimensions of $\mathbf{U}_{t}$ to get a retrieval key $\mathbf{R}_{t} \in \RR^{H_r \times F_g}$ as a rotation-invariant descriptor of the local map for fast global descriptor retrieval.
We employ the mean of ring-wise feature cosine distance as the comparative metric and select the closest historical frame $C_k$ as a loop candidate.
Then the rough relative rotation between the local map centered at the current and candidate loop frame can be estimated by shifting along the sector dimension of $\mathbf{U}_{t}$ and taking the shift number $n^{*}$ with the minimum bin-wise feature cosine distance $d_c^{*}$, which is calculated as the mean of $H_r \cdot H_s$ bins.
If $d_c^{*} < d_{\text{mc}}$, where $d_{\text{mc}}$ is the cosine distance threshold, we accept the loop candidate and use the relative rotation around $z$-axis by $\frac{2n^{*}\pi}{H_s}$ as the rotation part of the semi-metric localization.

Furthermore, to deal with the deficiency in translational invariance, we follow Scan Context++~\cite{kim2022tro} to use a polar context augmentation to enable the loop with large relative translation at the revisit. 
Specifically, for each scan position $\vv{t}_{WC_t}$, we generate $2V_a$ local map context descriptors at virtual positions ${\vv{\eta}_t^j}$ shifted from $\vv{t}_{WC_t}$ by increments of equal length $d_v$ along a direction perpendicular to the motion path $\delta \vv{t} = \vv{t}_{WC_t} - \vv{t}_{WC_{t-1}}$, results in virtual scan positions: 
\begin{equation}
  \vv{\eta}_t^j = \vv{t}_{WC_t} +  \frac{ j d_v}{\left\|\delta \vv{t}\right\|_2}\m{R}_z\left(90^\circ\right)\delta \vv{t}, ~~~  j=-V_a,\ldots,V_a,
\end{equation}
where $\m{R}_z\left(90^\circ\right)$ denotes the rotation around $z$-axis by $90^\circ$. 
%
%
We use these descriptors to query the most similar historical descriptors.
We use the one resulting in the minimum cosine distance and take $\vv{\eta}_t - \vv{t}_{WC_t}$ as the translation part of semi-metric localization results. 
We refer to Kim~\etal~\cite{kim2022tro} for more details regarding retrieval and augmentation.

When a candidate loop is detected, we register the current scan $\mathcal{P}_r$ to the local map $\mathcal{M}_l$ centered at the sensor position $\vv{t}_{WC_{k}}$ of the loop frame $C_{k}$ as described in \secref{sec:odometry}. 
For scan-to-map registration of local loops, we initialize the transformation with the identity matrix. 
For global loops, we use the results obtained from semi-metric localization as the initial guess for the transformation.
If the registration succeeds, we take the transformation $\mq{T}_{C_{k}C_{t}}$ resulting from the registration as the loop closure edge connecting node $C_{k}$ and $C_{t}$ in the pose graph $\mathcal{G}$. 
Otherwise, the candidate loop will be rejected.

\subsection{Globally Consistent Implicit Neural Map Adjustment}
\label{sec:pgo}

Once a loop is detected and verified by scan-to-map registration, we conduct pose graph optimization taking the current poses as the initial guess and using constant information matrices. 
To avoid the almost redundant optimization which would take up significant time, we disable the loop detection for $F_{\text{loop}}$ timesteps after conducting a pose graph optimization.
After performing pose graph optimization, the neural points in the global map $\mathcal{M}$ will move along with their associated frames. 
We apply a transformation to both the position and orientation of each neural point, as detailed in Sec.~\ref{sec:pin_map}. 
The transformation is determined according to the associated sensor frame $C_{t^m_i}$ of the neural point as the mean of its creation and last update timestep $t^m_i = \lfloor (t^c_i+t^u_i)/2 \rfloor$.
We denote the pose difference of frame $C_{t}$ before and after the pose graph optimization as $\delta \mq{T}_{t} = \mq{T}^{-1}_{WC_t^\prime}\mq{T}^{}_{WC_t}$ and the quaternion for the rotation part of $\delta \mq{T}_{t}$ as $\delta \vv{q}_{t}$. We can update the position $\vv{x}_i$ and orientation $\vv{q}_i$ of the neural point as:
\begin{equation} 
  \vv{x}_i \leftarrow \delta \mq{T}_{t^m_i} \vv{x}_i, ~~~
  \vv{q}_i \leftarrow \delta \vv{q}_{t^m_i} \vv{q}_i.
\end{equation}

After transforming the neural points, we recreate the voxel hash map and reset the local map to ensure correct neural point indexing.
If there are multiple neural points in one voxel, we keep the one whose stability $\mu$ is higher to avoid redundant map memory consumption in the revisiting region. 
%
%
To enable incremental mapping after pose graph optimization, we also transform the samples in the training sample pool $\mathcal{D}_p$ according to the optimized poses of their associated frames.

\section{Extension to RGB-D or Metric-Semantic SLAM}
\label{sec:extension}
As a proof of concept, we show how we can extend our approach to RGB-D and metric-semantic SLAM by extending the environment model with color or semantic information.

If the input point cloud contains point-wise RGB values or semantic classification probabilities predicted by any off-the-shelf segmentation models~\cite{milioto2019iros}, we can additionally predict the RGB value $\vv{c} \in \mathbb{R}^3$ and the $L$-class semantic probability $\vv{v} \in \mathbb{R}^{L}$ at $\vv{p}$ by each nearby neural point $\vv{m}_j$ in the same way as the SDF value prediction:
\begin{align}
  \vv{c}_j = D_\theta^c\left(\vv{f}_j^c, \vv{d}_j\right),~~~ 
  \vv{v}_j = D_\theta^s\left(\vv{f}_j^s, \vv{d}_j\right),
\end{align}
where $D_\theta^c$ and $D_\theta^s$ are the globally shared color and semantic decoder, $\vv{f}_j^c$ and $\vv{f}_j^s$ are the latent feature of neural point $\vv{m}_j$ assigned for color and semantics.
We can interpolate the final RGB prediction $\vv{c} = C\left(\vv{p}\right)$ and semantic prediction $\vv{v} = V\left(\vv{p}\right)$ in the same way as \eqref{equ:interpolation}.
To train the color and semantic field, we set the color target values $\hat{\vv{c}}_j$ and semantic pseudo labels $\hat{\vv{v}}_j$ for the $N_s+1$ close-to-surface samples as the same value at the endpoint. 
We then add the L1 loss $\mathcal{L}_\text{col}$ for color regression with a weight $\lambda_c$ or the $L$-class cross-entropy loss $\mathcal{L}_\text{sem}$ for semantic segmentation with a weight $\lambda_s$ to \eqref{equ:trainloss} using the close-to-surface training samples:
\begin{align}
  \mathcal{L}_\text{col}&=\frac{1}{N_c} \sum_{i=1}^{N_c} \left\| \hat{\vv{c}}_i - \vv{c}_i \right\|_1,\\
  \mathcal{L}_\text{sem}&=-\frac{1}{N_c} \sum_{i=1}^{N_c}\sum_{j=1}^L p^j_i \log \hat{p}^j_i,
\end{align}
where $\hat{\vv{c}}_i$ and $\vv{c}_i=C(\vv{u_W^i})$ are the color label and prediction while $p^j_i$ is the semantic prediction probability for class $j$ and $\hat{p}^j_i$ is the one-hot probability for the semantic pseudo label. 

For odometry estimation explained in \secref{sec:odometry}, with the point-wise color available, we add the photometric term into the optimization with a weight of $w_c$ to provide more constraints. 
In this case, we additionally minimize the photometric difference at the transformed position $\vv{p}_i^\prime$ between the observed value $\vv{c}_i$ and the predicted value $C\left(\vv{p}_i^\prime\right)$ from the PIN map as:
\begin{equation}
  \vv{\xi}^* = \underset{\vv{\xi}}{\operatorname{argmin}}\sum_{\vv{p}_{i} \in \mathcal{P}_r} S\left(\vv{p}_i^\prime\right)^2 + w_c \left\|C\left(\vv{p}_i^\prime\right) - \vv{c}_i\right\|_2^2.
\end{equation}
The Jacobian for the photometric term $\m{J}_i^c$ can be calculated likewise the geometric term as \eqref{equ:jacobian}, by replacing the SDF gradient $\vv{g}$ with the color gradient defined by $\nabla C\left(\vv{p}_i^\prime\right)$:
\begin{equation}
  \label{equ:jacobian2}
  \m{J}_i^c=\left[\nabla C\left(\vv{p}_i^\prime\right)^{\tr}, \big(\vv{p}_i^\prime \times \nabla C \left(\vv{p}_i^\prime \right)\big)^{\tr}\right].
\end{equation}
We then append $\m{J}_i^c$ to Jacobian matrix $\m{J}$ for the iterative optimization as \eqref{equ:lm}. 

During or after the mapping, we can colorize the mesh reconstructed from the SDF by querying the color or semantic value at the position of the mesh vertices.

\section{Experimental Evaluation}
\label{sec:exp}

%
The main focus of this work is a LiDAR SLAM system for building globally consistent maps using a point-based implicit neural map representation.

%
We present our experiments to show the capabilities of our method. The results of our experiments also support our key claims, which are:
(i) Our SLAM system achieves localization accuracy better or on par with state-of-the-art LiDAR odometry/SLAM approaches and is more accurate than recent implicit neural SLAM methods on various datasets using different range sensors. 
(ii) Our method can conduct large-scale globally consistent mapping with loop closure thanks to the elastic neural point representation.
(iii) Our map representation is more compact than the previous counterparts and can be used to reconstruct accurate and complete meshes at an arbitrary resolution.
(iv) Our correspondence-free scan-to-implicit map registration and the efficient neural point indexing by voxel hashing enable our algorithm to run at the sensor frame rate on a single NVIDIA A4000 GPU. 
%

\subsection{Experimental Setup}

\subsubsection{Datasets}
We extensively test our method on various datasets summarized in \tabref{tab:datasets}. 
For driving scenes, we test on KITTI, MulRAN and our self-collected dataset called IPB-car.
KITTI odometry dataset~\cite{geiger2012cvpr} comprises 22 sequences of LiDAR scans acquired by a Velodyne HDL64 LiDAR mounted on a car driving through Karlsruhe, Germany. 
Reference poses are available on sequences \text{00-10}.
MulRAN dataset~\cite{kim2020icra} is collected using an Ouster OS1-64 LiDAR mounted on a car traversing the roads of Daejeon, South Korea. 
Note that for the MulRAN dataset, the LiDAR is blocked by the Radar sensor, resulting in a smaller FoV than KITTI.
%
For KITTI and MulRAN, the poses measured by GNSS-INS are regarded as the reference for evaluation.
We additionally collected another challenging robot car dataset in Bonn, Germany with an OS1-64 LiDAR in 2020 and an OS1-128 LiDAR in 2023. 
We generate the reference poses by fusing GNSS-INS, LiDAR odometry \cite{vizzo2023ral}, loop closure constraints, and scan-to-map constraints between OS1-128 and a global map obtained by a geo-referenced terrestrial laser scanner (TLS) in a factor graph.

To further test our approach on handheld LiDARs which have a less constant motion profile, we adopt the Newer College and Hilti-21 dataset.
Newer College dataset~\cite{ramezani2020iros} contains two longer sequences acquired by a handheld OS1-64 LiDAR and a couple of shorter sequences collected by an OS0-128 LiDAR on the campus of Oxford University, UK.
The reference poses are obtained by aligning each scan to the survey grade point cloud map measured by TLS.
We also take the survey grade map as the reference model for mapping quality evaluation.
The Hilti 2021 SLAM challenge dataset~\cite{helmberger2022ral-hilti} comprises indoor sequences of offices, labs, and basements, as well as outdoor sequences of construction sites. 
These sequences were captured using a handheld OS0-64 LiDAR.
The reference trajectories are measured by either a total station tracking system or a motion capture system.
%

%

Aside from LiDAR datasets, we adopt the Replica synthetic RGB-D dataset~\cite{straub2019arxiv-replica}, which is a popular benchmark for recent neural RGB-D SLAM methods to show that our approach can also achieve precise pose estimation using RGB-D images. 

\begin{table}
  \centering
  \caption{Characteristic of the datasets used for evaluation}
  \resizebox{0.485\textwidth}{!}{
  \begin{tabular}{lllcc}
  \toprule
  dataset & sensor & scenario & \# seqs. & \# frames \\
  \midrule
  KITTI~\cite{geiger2012cvpr} & 64-beam LiDAR & outdoor, car & 22 & 43k\\
  MulRAN~\cite{kim2020icra} & 64-beam LiDAR & outdoor, car & 9 & 64k\\
  IPB-Car & 64/128-beam LiDAR & outdoor, car & 4 & 43k\\
  Newer College~\cite{ramezani2020iros} & 64/128-beam LiDAR & outdoor/indoor, handheld & 7 & 53k \\
  Hilti-21~\cite{helmberger2022ral-hilti} & 64-beam LiDAR &  outdoor/indoor, handheld & 6 & 15k \\
  Replica~\cite{straub2019arxiv-replica} & synthetic RGB-D & indoor, handheld & 8 & 16k \\
  \bottomrule
  \end{tabular}
  }
  \label{tab:datasets}
  \vspace{-2pt}
  \end{table}

\subsubsection{Parameters and Implementation Details}
%

\begin{table}
  \centering
  \caption{Hyperparameters of our approach}
  \resizebox{0.485\textwidth}{!}{
  \begin{tabular}{c|lll}
  \toprule
  type & symbol & value & description \\
  \midrule
  \multirow{11}{*}{PIN map} & $v_p$ & $0.005r_{\text{max}}$ & voxel hashing map resolution \\
  & $F_g$ & $8$ & dimensions of neural point latent features \\
  & $M_{\text{mlp}}, N_{\text{mlp}}$  & $2,64$ & MLP level and neuron count\\
  & $N_n$ & $5$ & neighborhood voxel count on each axis \\
  & $K$ & $6$ & neighborhood neural point number \\
  & $\sigma_{t}$ & $0.001r_{\text{max}}$ & sigmoid function scale factor \\
  & $\epsilon$ & $0.002r_{\text{max}}$ & perturbation step for numerical gradient \\
  & $\lambda_{e}$, $\lambda_{c}$, $\lambda_{s}$ & $0.5,0.5,1.0$ & weight for Ekional, color, and semantic loss \\
  & $r_l$  & $1.05r_{\text{max}}$ & radius of local map \\
  & $d_l$  & $4r_l$ & travel distance threshold of local map \\
  & $\gamma_d$, $\gamma_{\mu}$ & $0.008r_{\text{max}}, 4.0$ & dynamic filtering threshold for SDF and stability \\
  \midrule
  \multirow{6}{*}{sampling} & 
  $v_m$ & $0.001r_{\text{max}}$ & downsample voxel size for mapping \\
  & $\sigma_s$ & $0.003r_{\text{max}}$ & close-to-surface sampling standard deviation \\
  & $\zeta_{\text{min}}$ & 0.3 & minimum sample depth ratio \\
  & $d_b$ & $4\sigma_s$ & maximum sample range behind the surface \\
  & $N_s, N_f, N_b$ & $4, 2, 1$ & surface, front, and behind free space sample count\\
  & $N_p$ & $2 \times 10^7$ & maximum sample count in training sample pool \\
  \midrule
  \multirow{3}{*}{odometry} &
  $v_r$ & $0.0075r_{\text{max}}$ & downsample voxel size for registration\\
  & $w_c$ & $0.01$ & photometric tracking weight \\
  & $\kappa_r, \kappa_g$ & $0.005r_{\text{max}}, 0.1$ & GM kernel scale of residual and gradient anomaly \\
  \midrule
  \multirow{2}{*}{BA} & $F_{\text{ba}}$ & $20$ & implicit bundle adjustment frequency \\
  & $N_{\text{ba}}$ & $50$ & count of poses optimized during bundle adjustment \\
  \midrule
  \multirow{6}{*}{loop} & 
  $F_{\text{loop}}$ & $20$ & PGO frequency \\
  & $d_{\text{loop}}$ & $0.025r_{\text{max}}$ & distance threshold for local loop \\ 
  & $H_r, H_s$ & $20, 60$ & ring and sector resolution of polar descriptor \\
  & $d_{\text{mc}}$ & $0.3$ & cosine distance threshold for global loop \\ 
  & $V_a$ & $6$ & polar descriptor augmentation count  \\ 
  & $d_v$ & $0.02r_{\text{max}}$ & polar descriptor augmentation translation step   \\ 
  \bottomrule
  \end{tabular}
  }
  \label{tab:parameter}
  \vspace{-6pt}
\end{table}

We list the parameter setting of our approach in \tabref{tab:parameter}.
All the length-based parameters used in our method are set adaptively according to the maximum used measurement range $r_{\text{max}}$ of the sensor. 
The maximum range is $r_{\text{max}}=80\,\si{\metre}$ for KITTI, MulRAN, and self-recorded IPB-Car datasets and $r_{\text{max}}=60\,\si{\metre}$ for Newer College, and Hilti-21 datasets. 
For RGB-D datasets, we have $r_{\text{max}}=8\,\si{\metre}$. 
Our model is implemented mainly in PyTorch~\cite{paszke2019neurips}. 
For the online training of PIN map, we use Adam optimizer~\cite{kingma2015iclr} with a learning rate of $0.01$ and a batch size of $16384$ for $15$ iterations.  
Note that we train for $600$ iterations at the first frame for the map initialization.
To improve training efficiency, we only use 1/10\textsuperscript{th} of all training samples to calculate the numerical gradient for the Ekional loss $\mathcal{L}_{\text{eik}}$.
For local bundle adjustment, the poses are represented as Lie algebra tensors with PyPose~\cite{wang2023cvpr-pypose}.
We use a learning rate of $0.01$ for the neural point features, $0.0001$ for the poses, and a batch size of $16384$ for $80$ iterations. 
For pose graph optimization, we employ GTSAM~\cite{dellaert2012factor} and optimize the pose graph using Levenberg-Marquardt optimization with a maximum of $50$ iterations.





\begin{table*}[t]
  \caption{LiDAR odometry performance comparison on \textit{KITTI} LiDAR dataset with motion compensated point cloud with the average relative translational drifting error (\%). The numbers with \dag ~are the results of the training set for learning-based methods and are not included in the ranking. - indicates the number is not reported. We highlight the best results in \textbf{bold} and the second best in \underline{underscored}. For the results of our method PIN-LO, we report the mean and standard deviation (in brackets) calculated from 10 runs with different random seeds. }
  \centering
  
  \resizebox{0.985\textwidth}{!}{
  \begin{tabular}{cc|ccccccccccc|cc}
    \toprule
    Method & Type & \text{00} & \text{01} & \text{02} & \text{03} & \text{04} & \text{05} & \text{06} & \text{07} & \text{08} & \text{09} & \text{10} & 
    \textbf{Avg.} & \text{11-21} \\  
    \midrule
    LeGO-LOAM~\cite{shan2018iros} & feature points & 2.17 & 13.4 & 2.17 & 2.34 & 1.27 & 1.28 & 1.06 & 1.12 & 1.99 & 1.97 & 2.21 & 2.49 & -  \\
    F-LOAM~\cite{wang2021iros-fflo} & feature points   & 0.78 & 1.43 &	0.92 & 0.86 &	0.71 & 0.57 &	0.65 & 0.63 &	1.12 & 0.77	& 0.79 & 0.84 & 0.72 \\
    MULLS~\cite{pan2021icra-mvls}  & feature points   & 0.56 & \underline{0.64}	& 0.55 & 0.71 &	0.41 & \underline{0.30} &	0.30 & 0.38 &	\bf{0.78} & \bf{0.48} &	0.59 & 0.52 & 0.65 \\
    VG-ICP~\cite{koide2021icra} & dense points   & 2.16 & 2.38	& 0.99 & 0.67	& 0.55 & 0.45	& \bf{0.24} & 0.99 &	1.74 & 0.95	& 0.95 & 1.10 &   -  \\
    CT-ICP~\cite{dellenbach2022icra} & dense points  & \bf{0.49} & 0.76 & \underline{0.52} & 0.72 & 0.39 & \bf{0.25} & 0.27 & \bf{0.31} & 0.81 & \underline{0.49} & \bf{0.48} & \bf{0.50} & \bf{0.59} \\
    KISS-ICP~\cite{vizzo2023ral} & dense points  & 0.52 & \bf{0.63} & \bf{0.51} & \underline{0.66} &	0.36 & 0.31 &	\underline{0.26} & \underline{0.33} &	0.82 & 0.51 &	0.56 & \bf{0.50} & \underline{0.61} \\
    SuMa-LO~\cite{behley2018rss} & surfels & 0.73	& 1.71 & 1.06 &	\underline{0.66} & 0.38 &	0.50 & 0.42 &	0.39 & 1.02 &	0.48 & 0.71	& 0.73 & 1.39 \\
    Litamin-LO~\cite{masashi2021icra} & normal distribution &  0.78 & 2.10 & 0.95 & 0.96 & 1.05 & 0.55 & 0.55 & 0.48 & 1.01 & 0.69 & 0.80 & 0.88 & -\\
    IMLS-SLAM~\cite{deschaud2018icra} & implicit model & \underline{0.50} & 0.82 & 0.53 & 0.68 & \underline{0.33} & 0.32 & 0.33 & \underline{0.33} & \underline{0.80} & 0.55 & 0.53 & 0.52 & 0.69 \\
    Puma~\cite{vizzo2021icra}  & mesh   & 1.46 & 3.38 &	1.86 & 1.60 &	1.63 & 1.20 &	0.88 & 0.72 &	1.44 & 1.51 &	1.38 & 1.55 &   -  \\
    SLAMesh~\cite{ruan2023icra} & mesh  & 0.77 & 1.25 &	0.77 & \bf{0.64} &	0.50 & 0.52 &	0.53 & 0.36 &	0.87 & 0.57 &	0.65 & 0.68 &   -  \\
    \midrule
    LONet~\cite{li2019cvpr-lonet} & supervised & 1.47$^\dag$ & 1.36$^\dag$ & 1.52$^\dag$ & 1.03$^\dag$ & 0.51$^\dag$ & 1.04$^\dag$ & 0.71$^\dag$ & 1.70 & 2.12 & 1.37 & 1.80 & 1.33 & -\\
    PWCLONet~\cite{wang2021cvpr-pwclo} & supervised & 0.78$^\dag$ & 0.67$^\dag$	& 0.86$^\dag$ & 0.76$^\dag$	& 0.37$^\dag$ & 0.45$^\dag$	& 0.27$^\dag$ & 0.60 &	1.26 & 0.79	& 1.69 & 0.77 &  -  \\
    ELONet~\cite{wang2022pami} & supervised & 0.83$^\dag$	& 0.55$^\dag$	& 0.71$^\dag$	& 0.49$^\dag$ &	0.22$^\dag$ &	0.34$^\dag$ & 0.36$^\dag$ &	0.46	& 1.14 &	0.78 &	0.80 &  0.61 & 1.92\\
    Nerf-LOAM~\cite{deng2023iccv} & neural implicit & 1.34 & 2.07 & -    & 2.22 &	1.74 & 1.40 &  -   & 1.00 &  -	   & 1.63 &	2.08 & 1.69 &   -  \\
    \midrule
    \multirow{2}{*}{PIN-LO} & \multirow{2}{*}{neural implicit} & 0.55 &	0.68	& 0.54 &	0.76 &	\bf{0.22} &	\underline{0.30} &	0.35 &	0.34 &	\underline{0.80}	& 0.54	& \underline{0.50}	& \underline{0.51}  & 0.64\\
     & & (\textpm0.02) &	(\textpm0.13) & (\textpm0.02) &	(\textpm0.02) &	(\textpm0.02) &	(\textpm0.01)	& (\textpm0.01) &	(\textpm0.02)	& (\textpm0.02)	 & (\textpm0.06)	& (\textpm0.05) &	(\textpm0.02) & - \\
    \bottomrule
  \end{tabular}
  } 
  \vspace{-2pt}
  \label{tab:kitti_odom}
\end{table*}

\begin{table*}[t]
  \caption{SLAM performance comparison (ATE RMSE [m]) on \textit{KITTI} LiDAR dataset with motion compensated point cloud. $*$ indicates the sequence is with loops, Avg.* denotes the average metric on sequences with loops. We highlight the best results in \textbf{bold} and the second best in \underline{underscored}.  $\ddagger$ indicates the method conducts offline pose graph optimization. For the results of our method PIN-LO and PIN-SLAM, we report the mean and standard deviation (in brackets) calculated from 10 runs with different random seeds.}
  \centering
  \resizebox{0.85\textwidth}{!}{
  \begin{tabular}{c|ccccccccccc|cc}
    \toprule
    Method & \text{00}* & \text{01} & \text{02}* & \text{03} & \text{04} & \text{05}* & \text{06}* & \text{07}* & \text{08}* & \text{09}* & \text{10} & \textbf{Avg.*} & Avg.         \\
    \midrule
    SuMa~\cite{behley2018rss} & 1.1 & 14.6 & 8.0 &	1.0 & \underline{0.3} &	0.7 & 0.6 &	1.1	& 3.7 &	1.2	& 1.4	& 2.3	& 3.1 \\
    MULLS~\cite{pan2021icra-mvls}   & 1.1 & \bf{1.9}	& 5.4 & 0.7 &	0.9 & 1.0 &	\underline{0.3} & \underline{0.4} &	2.9 & 2.1 &	1.1 & 1.9 & 1.6   \\
    Litamin2~\cite{masashi2021icra} & 1.3 & 15.9 & 3.2 & 0.8 & 0.7 & 0.6 & 0.8 & 0.5 & \textbf{2.1} & 2.1 & \underline{1.0} & \underline{1.5} & 2.4 \\
    SC-LeGO-LOAM~\cite{shan2018iros,kim2018iros} & 2.3 & 19.7 &	5.3 &	1.6 &	0.4 &	1.2 &	1.0 & 1.5 &	5.9	& 2.0	& 1.7 &	2.7 & 5.3 \\
    HLBA~\cite{liu2023ral}$^\ddagger$   & \bf{0.8} & \bf{1.9} & 5.1 & \underline{0.6} & 0.8 & \underline{0.4} & \bf{0.2} & \bf{0.3} & 2.7 & 1.3 & 1.1 & \underline{1.5} & 1.4  \\
    SC-F-LOAM~\cite{wang2021iros-fflo,kim2018iros}$^\ddagger$ & 1.3 &	4.7 &	3.3 &	0.7	& \underline{0.3} &	1.2 &	0.4 &	0.5 &	3.0 &	1.3 &	1.6	& 1.6	& 1.7 \\
    SC-KISS-ICP~\cite{vizzo2023ral,kim2018iros}$^\ddagger$ & \underline{1.0}	& \underline{3.7}	& \textbf{1.9}	& \textbf{0.4}	& \underline{0.3}	& \underline{0.3}	& \underline{0.3}	& \textbf{0.3}	& \underline{2.2}	& \textbf{1.0} & \textbf{0.8}	& \bf{1.0}	& \bf{1.1} \\
    \midrule
    \multirow{2}{*}{PIN-LO} & 5.6 &	4.3 & 9.3 &	0.7 &	\textbf{0.1} &	1.7 &	0.5 &	0.5 &	3.0 &	1.8 &	\textbf{0.8} & 3.2 & 2.6 \\
    &  (\textpm0.3) &	(\textpm1.7) & (\textpm0.7) &	(\textpm0.1) &	(\textpm0.0) &	(\textpm0.1)	& (\textpm0.0) &	(\textpm0.0)	& (\textpm0.2)	 & (\textpm0.4)	& (\textpm0.1) &	(\textpm0.3) &	(\textpm0.2) \\
    \midrule
    \multirow{2}{*}{PIN-SLAM} & \textbf{0.8} & 4.3	& \underline{2.1} &	0.7 &	\textbf{0.1} &	\textbf{0.3} &	0.4 &	\textbf{0.3}	& \textbf{2.1} &	\underline{1.2} &	\textbf{0.8} &	\bf{1.0} & \underline{1.2}\\
    &  (\textpm0.1) &	(\textpm1.7) & (\textpm0.4) &	(\textpm0.1) &	(\textpm0.0) &	(\textpm0.0)	& (\textpm0.0) &	(\textpm0.0)	& (\textpm0.4)	 & (\textpm0.1)	& (\textpm0.1) &	(\textpm0.1) &	(\textpm0.2) \\
    \bottomrule
  \end{tabular}
  }
  \vspace{-4pt}
  \label{tab:kitti_slam}
\end{table*}

\begin{figure*}[!htbp]
  \centering  
  \includegraphics[width=0.92\linewidth]{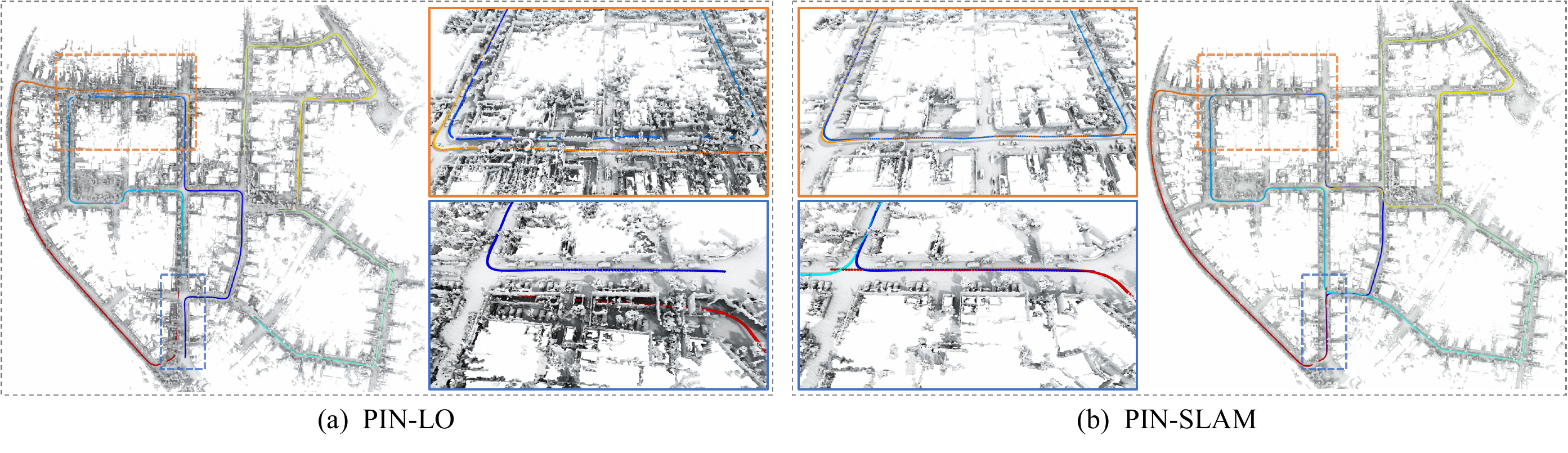}
  \setlength{\abovecaptionskip}{0pt}
  \caption{Comparison of (a) the locally consistent mesh with duplicated structures reconstructed by PIN LiDAR odometry, and (b) the globally consistent mesh reconstructed by PIN-SLAM built on KITTI sequence \text{00} after loop closure corrections. The estimated trajectories are overlaid on the map and colorized according to the timestamp. Details of two revisited regions are highlighted in the boxes.}
  \label{fig:kitti_00_compare}
  \vspace{-8pt}
\end{figure*}

\subsection{Localization Accuracy Evaluation}

In this section, we compare the pose estimation performance of PIN-SLAM with state-of-the-art odometry/SLAM systems.

\subsubsection{LiDAR Odometry Evaluation on KITTI}
We first evaluate the pure odometry accuracy of our approach on the competitive KITTI odometry benchmark~\cite{geiger2012cvpr}. 
We denote the LiDAR odometry version of our approach as PIN-LO, which disables the loop closure detection correction module of PIN-SLAM.
Since the LiDAR scans in KITTI odometry dataset are already deskewed, we disable the motion compensation for our method.
We compare PIN LiDAR odometry against various LiDAR odometry systems using different map representations such as feature points~\cite{shan2018iros, wang2021iros-fflo, pan2021icra-mvls}, denser voxel-downsampling points~\cite{koide2021icra, dellenbach2022icra, vizzo2023ral}, normal distribution transformation~\cite{masashi2021icra}, surfels~\cite{behley2018rss}, IMLS model~\cite{deschaud2018icra}, and triangle meshes~\cite{vizzo2021icra, ruan2023icra}.
Since our approach is based on online learning of a neural implicit map, we additionally compare our method against the learning-based LiDAR odometry systems, including those using supervised learning~\cite{li2019cvpr-lonet, wang2021cvpr-pwclo, wang2022pami} and Nerf-LOAM~\cite{deng2023iccv} which also utilizes a neural implicit map. 
Note that those systems based on supervised learning are trained on sequences \text{00-06} and tested on sequences \text{07-10} while our method and Nerf-LOAM do not rely on any pre-training.
We use the average relative translational error (ARTE)~\cite{geiger2012cvpr} as the metric for odometry drift evaluation.
Due to the non-deterministic nature of online training in our method, the results may have slight variations with different random seeds. 
Therefore, we opt to report the mean and standard deviation of the metrics calculated from 10 different runs with varying random seeds on the KITTI dataset. 
This allows a fair comparison with other methods and provides a better understanding of the stochastic nature of PIN-SLAM. 
As shown in \tabref{tab:kitti_odom}, PIN LiDAR odometry outperforms all the learning-based methods and most of the non-learning-based methods.
It attains an average translation error of $0.51\%$ with a standard deviation of $0.02\%$, which is on par with the two open-source state-of-the-art LiDAR odometry KISS-ICP~\cite{vizzo2023ral} and CT-ICP~\cite{dellenbach2022icra} using a voxel-downsampled point cloud map.

\begin{table*}[t]
  \centering
  \caption{SLAM performance comparison (ATE RMSE [m]) of our method vs. the state-of-the-art LiDAR odometry (above) and LiDAR SLAM methods (below) on \textit{MulRAN} LiDAR dataset. All sequences are with loops. The best result is in \textbf{bold} and the second best is \underline{underscored}. $\ddagger$ indicates the method conducts offline pose graph optimization.} 
  \label{tab:ate_mulran}
  \resizebox{0.74\textwidth}{!}{
      \begin{tabular}{c|ccccccccc|c}
          \toprule
          Method   & \text{KA1}     & \text{KA2}   & \text{KA3} & \text{DC1} & \text{DC2} & \text{DC3}  & \text{RS1}   & \text{RS2}     & \text{RS3}  & \textbf{Avg.}   \\
          \midrule
          F-LOAM~\cite{wang2021iros-fflo} & 49.16	& 35.93 &	38.36 &	36.89 &	26.72 &	31.97 &	91.85 &	104.25 &	119.79 & 59.44 \\ 
          KISS-ICP~\cite{vizzo2023ral} & 20.81 & 13.52 & 14.85 & 20.26 & 12.79 & 13.17 &  14.35 & 32.74 & 75.14 & 24.18 \\
          \midrule
          SuMa~\cite{behley2018rss} & 10.45 & 9.33 & 10.33 & 13.00 & 86.04 & 12.03 & 114.89 & 818.74 & 367.41 & 160.25 \\ 
          MULLS~\cite{pan2021icra-mvls} & 19.72 & 15.26 & 6.93 & 14.95 & 11.31 & 6.00 & 41.25  & 45.05 & 44.18 &  22.74 \\
          SC-LeGO-LOAM~\cite{shan2018iros,kim2018iros} & 5.45 & 5.49 & 5.70 & 6.95 & 5.49 & 6.29 & 19.05 & 16.04 & 30.91 & 11.26 \\
          HLBA~\cite{liu2023ral}$^\ddagger$ & 3.36 & 3.75 & 3.53 & \bf{5.20} & \underline{3.22} & \underline{2.54} & 8.92 & \bf{7.94} & 10.26 & 5.71 \\
          SC-F-LOAM~\cite{wang2021iros-fflo,kim2018iros}$^\ddagger$ & 4.74 &	4.70	& 4.32 &	9.67 &	5.57 &	3.98 &	17.72 &	22.42 &	24.07 &	10.80 \\ 
          SC-KISS-ICP~\cite{vizzo2023ral,kim2018iros}$^\ddagger$& \underline{3.33}	& \bf{2.80} &	\underline{2.65} &	6.41 &	3.42 &	\bf{2.13} &	\bf{6.59} &	\underline{9.45} &	\underline{8.97} &	\underline{5.08} \\ 
          \midrule
          PIN-LO & 29.07 & 24.74 & 23.05 & 22.07 & 12.94 & 23.42 & 45.81 & 54.41 & 44.41 &	31.10  \\
          %
          PIN-SLAM & \bf{2.25}	 & \underline{2.86} &	\bf{2.15} &	\underline{5.25} &	\bf{2.83} &	2.65 &	\underline{8.70}	& \bf{7.94}	 & \bf{6.00}	& \bf{4.51} \\
          \bottomrule
      \end{tabular}
  }
  \vspace{-9pt}
\end{table*}

\begin{table}[!t]
  \vspace{0.2cm}
  \centering
  \caption{SLAM performance comparison (ATE RMSE [m]) of the proposed method vs. the state-of-the-art LiDAR odometry (above) and LiDAR SLAM methods (below) on our self-collected \textit{IPB-Car} LiDAR dataset. $*$ means the sequence is with loops. The best result is in \textbf{bold} and the second best is \underline{underscored}.}
  \label{tab:ate_ipbcar}
  \resizebox{0.43\textwidth}{!}{
      \begin{tabular}{r|cccc|c}
          \toprule
          Method   & \text{2020-0}* & \text{2020-1}* & \text{2023-0} & \text{2023-1}* & \textbf{Avg.} \\
          \midrule
          F-LOAM~\cite{wang2021iros-fflo} & 11.75 & 39.92 & 92.48 & 52.70 & 49.21 \\
          KISS-ICP~\cite{vizzo2023ral} & 6.13 & \underline{15.66} & 93.70 & 22.91 & 34.60 \\
          \midrule
          SuMa~\cite{behley2018rss} & 7.15 & 117.91 & 100.12 & 57.66 & 70.71\\
          MULLS~\cite{pan2021icra-mvls} & 10.79 & 47.07 & \bf{78.10} & 68.63 & 51.15 \\
          \midrule
          PIN-LO & \underline{5.70}  & 17.42 & \underline{87.59} & \underline{21.44} & \underline{33.04} \\
          PIN-SLAM & \bf{3.51} & \bf{6.12} & \underline{87.59} & \bf{19.27} & \bf{29.12}\\
          \bottomrule
      \end{tabular}
  }
  \vspace{0.5cm}
  %
  \caption{SLAM performance comparison (ATE RMSE [m]) of our method vs. the state-of-the-art LiDAR odometry (above) and LiDAR SLAM methods (below) on \textit{Newer College} handheld LiDAR dataset. All sequences are with loops. The best result is \textbf{bold} and the second best is \underline{underscored}. \xmark ~denotes failure. - indicates the number is not reported and unavailable using the open-source code. } 
  \label{tab:ate_ncd}
  \resizebox{0.48\textwidth}{!}{
      \begin{tabular}{r|ccccccc|c}
          \toprule
          Method                                    & \text{01}     & \text{02}      & \text{quad\_e}         &  \text{math\_e}    & \text{ug\_e}  &  \text{cloister}     & \text{stairs}  & \textbf{Avg.} \\
          \midrule
          F-LOAM~\cite{wang2021iros-fflo} & 6.74 & \xmark & 0.40 &  0.26 &  \underline{0.09} & 7.69 & \xmark & 3.04  \\
          KISS-ICP~\cite{vizzo2023ral} & \underline{0.62} & 1.88 & \underline{0.10} & \bf{0.07} & 0.33 & 0.30 & \xmark & \underline{0.55} \\ 
          \midrule
          SuMa~\cite{behley2018rss} & 2.03 & 3.65 & 0.28 & 0.16 & \underline{0.09} & 0.20 & 1.85 & 1.18 \\
          MULLS~\cite{pan2021icra-mvls} &  2.51 & 8.39 & 0.12 & 0.35 & 0.86 & 0.41 & \xmark & 2.11 \\ 
          MD-SLAM~\cite{di2022iros} &  - & 1.74 & 0.25 & - & - & 0.36 & 0.34 &  -\\
          SC-LeGO-LOAM~\cite{shan2018iros,kim2018iros} & - & \underline{1.30} & \bf{0.09} & - & - & 0.20 & 3.20 & - \\ 
          \midrule
          PIN-LO & 2.08 & 5.32 & \bf{0.09} & \underline{0.09} & \bf{0.07} &  \underline{0.19} &  \underline{0.07} & 1.13 \\
          PIN-SLAM & \bf{0.43} & \bf{0.31} & \bf{0.09} & \underline{0.09} & \bf{0.07} &\bf{0.15} & \bf{0.06} & \bf{0.17} \\
          \bottomrule
      \end{tabular}
  }
  \vspace{0.5cm}
  %
  \centering
  \caption{SLAM performance comparison (ATE RMSE [m]) of the proposed method vs. the state-of-the-art methods on \textit{Hilti-21}  LiDAR dataset. Best result is \textbf{bold} and the second best is \underline{underscored}. }
  \label{tab:ate_hilti}
  \resizebox{0.48\textwidth}{!}{
      \begin{tabular}{r|cccccc|c}
        \toprule
          Method  & \text{rpg} & \text{lab}  & \text{base1}  & \text{base4} & \text{cons2} & \text{camp2} & \textbf{Avg.}  \\
          \midrule
          F-LOAM~\cite{wang2021iros-fflo} & 2.78 & 0.18  & 0.91 & 0.29 & 11.52 & 8.95 & 4.10 \\
          KISS-ICP~\cite{vizzo2023ral} & \underline{0.22} & 0.07 & 0.32 & \underline{0.11} & 0.84 & 1.98 & 0.58 \\
          HDLGraph-SLAM~\cite{koide2019ijars} & 0.35 & \underline{0.05} & \bf{0.28} & 0.37 & \underline{0.74} & \underline{0.35} & \underline{0.36} \\
          \midrule
          PIN-SLAM & \bf{0.21} & \bf{0.04} & \underline{0.30} & \bf{0.08} & \bf{0.41} & \bf{0.11} & \bf{0.19}
          \\
          \bottomrule
      \end{tabular}
  }
  \vspace{-6pt}
\end{table}

Our method demonstrates superior performance compared to Nerf-LOAM~\cite{deng2023iccv}, the sole baseline using also an implicit neural map representation. 
This achievement is attributed to our better SDF training in the close-to-surface free space and the more robust point-to-SDF registration using Levenberg-Marquardt optimization combined with robust kernels instead of the gradient descent used by Nerf-LOAM.

Additionally, as shown in the last column of \tabref{tab:kitti_odom}, our method performs among the best LiDAR odometry/SLAM systems on the hidden sequences \text{11-21} of the KITTI odometry leaderboard and is the best-ranked learning-based method.

\subsubsection{LiDAR SLAM Evaluation}
%
We proceed to evaluate the full system of PIN-SLAM, including the loop closure correction module.
We test PIN-SLAM quantitatively and compare it against state-of-the-art LiDAR SLAM/odometry approaches on four public datasets and one self-recorded dataset collected by different robot platforms in various scenarios.
Instead of using the average relative translational error, we adopt the root mean square error (RMSE) of the absolute trajectory error (ATE)~\cite{zhang2018iros-evo} with Umeyama trajectory alignment as the localization accuracy metric since it can better reflect the global consistency of the estimated pose. 
For comparison, we focus on methods that either disclose their achieved ATE RMSE on our utilized datasets in their papers or release open-sourced code that supports our utilized datasets.

Firstly, on the KITTI sequence \text{00-10}, we compare our approach with four state-of-the-art LiDAR SLAM systems enabling loop closure correction, and HLBA~\cite{liu2023ral}, a postprocessing method using LiDAR bundle adjustment with the pose initial guess of MULLS~\cite{pan2021icra-mvls}.
In line with previous work~\cite{masashi2021icra,liu2023ral}, we report the accuracy at decimeter precision.
As shown in \tabref{tab:kitti_slam}, PIN-SLAM achieves the smallest average RMSE of $1.0\,\si{\metre}$ on the sequences with loops and $1.2\,\si{\metre}$ on all the eleven sequences with a standard deviation of $0.2\,\si{\metre}$ calculated from 10 runs with different random seeds.
Notably, PIN-SLAM even outperforms the post-processing approach~\cite{liu2023ral} while PIN-SLAM can run online.
In addition to SC-LeGO-LOAM, an open-source SLAM system combining LeGO-LOAM~\cite{shan2018iros} with scan context~\cite{kim2018iros}, we additionally implement SC-F-LOAM and SC-KISS-ICP, which perform offline pose graph optimization using the pose of more recent LiDAR odometry systems F-LOAM and KISS-ICP with the loop closures detected by scan context. 
We conduct a fine ICP registration between the query and retrieved point cloud to refine the loop transformation initial guess.
PIN-SLAM outperforms SC-F-LOAM and achieves comparable localization accuracy to SC-KISS-ICP.
Meanwhile, PIN-SLAM can maintain a continuous SDF map online for downstream tasks such as mesh reconstruction or path planning whereas the compared methods typically construct only a sparse point cloud map.
We also report the result of PIN LiDAR odometry to show the significant improvement of PIN-SLAM on trajectory global consistency, with the RMSE decreasing from $3.2\,\si{\metre}$ to $1.0\,\si{\metre}$ on sequences with loop closures. 
As shown in \figref{fig:kitti_00_compare}, PIN-SLAM manages to correct the drift of PIN LiDAR odometry and build a globally consistent map without duplicated structures.

Next, we test PIN-SLAM on longer and more challenging driving scenes, \ie, nine sequences of the MulRan dataset and four sequences of our self-recorded IPB-Car dataset.  
As shown in \tabref{tab:ate_mulran} and \tabref{tab:ate_ipbcar}, compared with the state-of-the-art LiDAR odometry, SLAM or offline optimized approaches, PIN-SLAM is able to correct the drift of PIN LiDAR odometry and achieves the best overall localization accuracy.

To further test our method on other motion profiles, we run experiments on Newer College and Hilti-21 datasets, which are mainly collected by a handheld LiDAR in both indoor and outdoor environments. 
We evaluate PIN-SLAM against existing LiDAR odometry/SLAM systems supporting handheld LiDARs.
As shown in \tabref{tab:ate_ncd} and \tabref{tab:ate_hilti}, PIN-SLAM demonstrates superior performance, yielding the smallest localization error on both datasets and surpassing the compared approaches by a large margin.
For the Newer College dataset, half of the compared methods fail on the challenging stairs sequence while our approach achieves the smallest RMSE of 6\,cm.
Note that the sequences in Hilti-21 dataset are relatively short and the LiDAR is mainly moving in a confined area without explicit loops. 
Therefore we do not distinguish an odometry and a SLAM method on Hilti-21.

\begin{table}[t]
  \caption{SLAM performance comparison (ATE RMSE [cm]) of the proposed method vs. the state-of-the-art methods on \textit{Replica} RGB-D dataset. Best result is \textbf{bold} and second best is \underline{underscored}. }
  \centering
  \resizebox{0.485\textwidth}{!}{
  \begin{tabular}{r|cccccccc|c}
    \toprule
    Method &  \text{r0} & \text{r1} & \text{r2} & \text{o0} & \text{o1} & \text{o2} & \text{o3} & \text{o4} & \textbf{Avg.} \\                                
    \midrule
    iMAP~\cite{sucar2021iccv} & 3.12 & 2.54 & 2.31 & 1.69 & 1.03 & 3.99 & 4.05 & 1.93 & 2.58 \\
    NICE-SLAM~\cite{zhu2022cvpr}  & 1.69 & 2.04 & 1.55 & 0.99 & 0.90 & 1.39 & 3.97 & 3.08 & 1.95 \\
    Vox-Fusion~\cite{yang2022ismar} & \underline{0.40} & 0.54 &	0.54 & 0.50 &	\underline{0.46} & 0.75 &	\underline{0.50} & \underline{0.60} &	0.54 \\
    Co-SLAM~\cite{wang2023cvpr-coslam} & 0.60 & 1.13 & 1.43 & 0.55 & 0.50 & \underline{0.46} & 1.40 & 0.77 & 0.86 \\
    ESLAM~\cite{johari2023cvpr} & 0.71	& 0.70 &	0.52 &	0.57 &0.55 &	0.58 &	0.72 &	0.63 & 0.63 \\
    Point-SLAM~\cite{sandstrom2023iccv} & 0.61 & \underline{0.41} & \underline{0.37} & \underline{0.38} & 0.48 & 0.54 & 0.72 & 0.63 & \underline{0.52} \\
    \midrule
    PIN-SLAM  & \bf{0.27} &	\bf{0.31} &	\bf{0.13} &	\bf{0.22} &	\bf{0.30} &	\bf{0.28} &	\bf{0.16} &	\bf{0.28} &	\bf{0.24} \\
    \midrule
    PIN-SLAM w/o BA & 0.41 & 0.36 &	0.25 & 0.36 &	0.33 & 0.34 &	0.38 &	0.73	& 0.40 \\ 
    PIN-SLAM w/o color & 0.38	& 0.40 & 0.16 &	0.39 & 0.27 &	0.30 & 0.52 &	0.40	& 0.35 \\
    \bottomrule
  \end{tabular}
  }
  \vspace{-10pt}
  \label{tab:rgbd_replica}
\end{table}

\begin{figure}[!t]
  \centering  
  \includegraphics[width=0.95\linewidth]{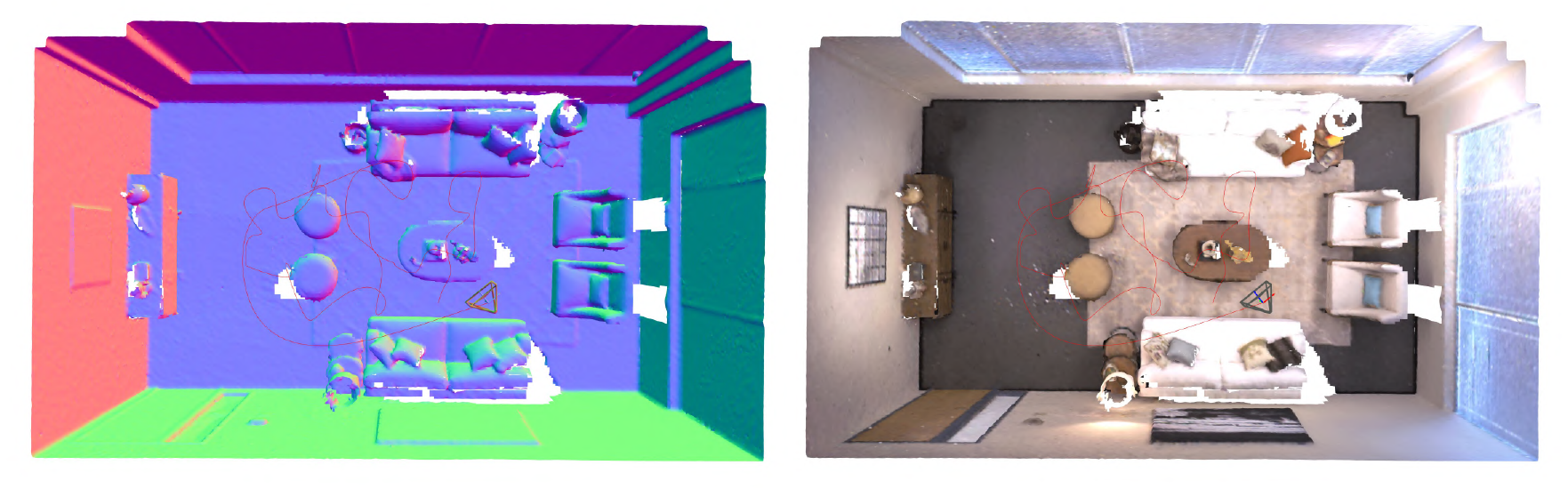}
  \setlength{\abovecaptionskip}{2pt}
  \caption{Qualitative results of PIN-SLAM on \text{Room0}  sequence of \textit{Replica} RGB-D dataset. We show the mesh colorized in the normal direction and the mesh colorized by RGB built by PIN-SLAM. The estimated trajectory is shown in red.}
  \label{fig:replica_quali}
  \vspace{-8pt}
\end{figure}

\subsubsection{RGB-D SLAM Evaluation}
We also conducted experiments to show that PIN-SLAM can also work well taking as input the RGB-D images. 
We compare the localization (camera pose tracking) accuracy of PIN-SLAM against six state-of-the-art neural implicit RGB-D SLAM approaches on all 8 sequences of the popular Replica dataset.
Since the camera is moving in a single room that is always covered by the local map, we disable the loop closure detection and correction.
As shown in \tabref{tab:rgbd_replica}, PIN-SLAM achieves the best camera tracking accuracy on average with an RMSE of only 2.4\,mm.
The superior performance demonstrates that, with precise depth measurements, our point-to-SDF registration achieves greater accuracy compared to the rendering-based camera tracking utilized in most of the neural implicit SLAM approaches under comparison.
As an ablation study, we show that the localization RMSE of PIN-SLAM decreases by 40\% and 31\% by conducting local bundle adjustment and using the color information during camera tracking.
Moreover, as shown in \figref{fig:replica_quali}, PIN-SLAM can reconstruct high-fidelity colorized mesh from the neural point map built by the RGB-D SLAM.

\subsubsection{Additional Challenging Scenarios}

\begin{figure}[!t]
  \centering  
  \includegraphics[width=0.97\linewidth]{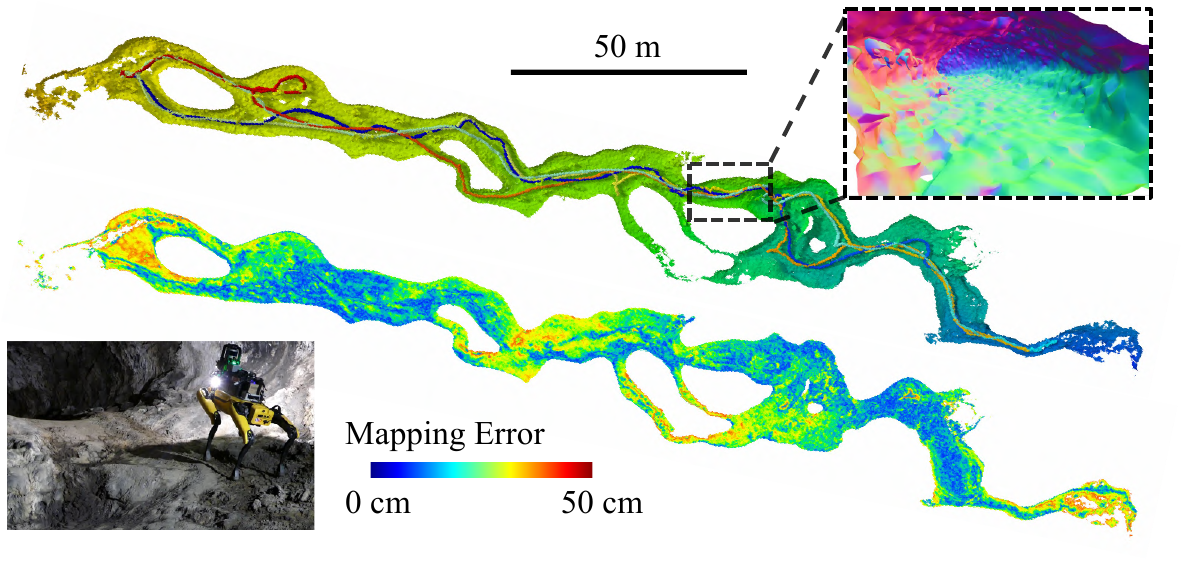}
  \setlength{\abovecaptionskip}{0pt}
  \caption{Qualitative results on the \textit{Nebula} dataset collected by a Spot1 robot with a 32-beam LiDAR moving back-and-forth in the Valentine Cave. We show the estimated trajectory and mesh built by PIN-SLAM on the top. We show the mapping error compared to the survey-grade map measured by a terrestrial laser scanner on the bottom.}
  \vspace{-6pt}
  \label{fig:nebula_quali}
\end{figure}

\begin{figure}[!t]
  \centering  
  \includegraphics[width=0.97\linewidth]{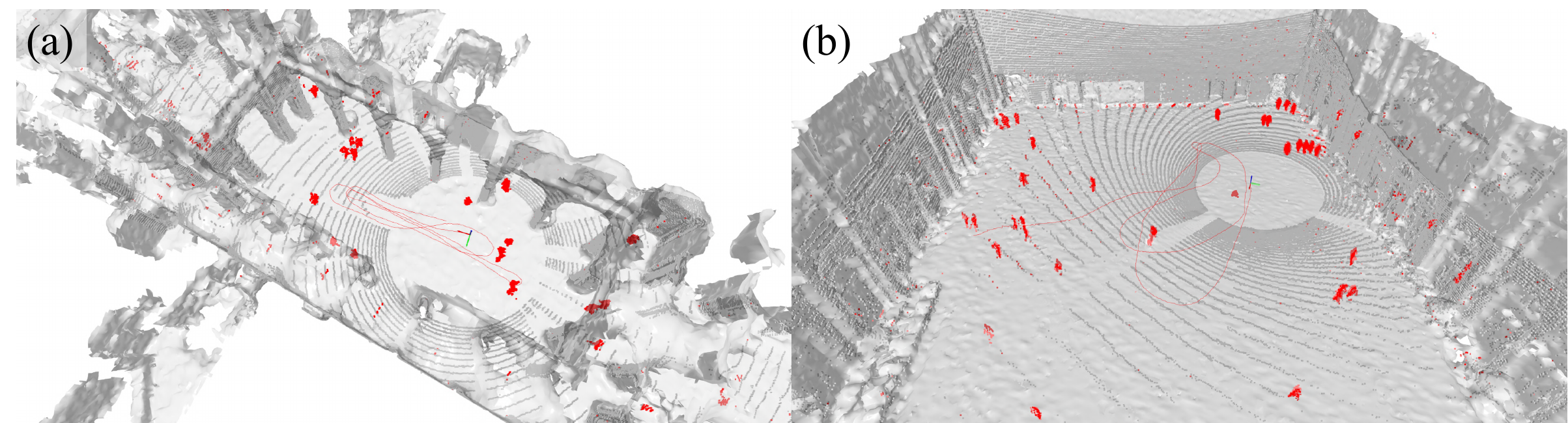}
  \setlength{\abovecaptionskip}{4pt}
  \caption{Qualitative results on the \text{hauptgebaeude} and \text{station} sequences of \textit{ETH DOALS} dataset. We show the lidar scan overlaid on the static mesh built by PIN-SLAM.
  The dynamic points (in red) are filtered from a LiDAR scan (in gray) online using the PIN map.}
  \vspace{2pt}
  \label{fig:eth_dynamic_quali}
\end{figure}

We show qualitative results on various challenging scenarios such as a cave tunnel from Nebula dataset~\cite{reinke2022ral-iros} as shown in \figref{fig:nebula_quali}. 
PIN-SLAM manages to build a globally consistent map using the data collected by a quadruped robot moving in the cave.
PIN-SLAM is also robust to highly dynamic scenes in ETH DOALS dataset~\cite{schmid2023ral-dynablox} as shown in \figref{fig:eth_dynamic_quali}. 
PIN-SLAM manages to filter the moving pedestrians and reconstruct a static mesh.
%

\begin{table}[!t]
	\centering
	\caption{Loop closure detection recall at Top-1 of the proposed method vs. the state-of-the-art non-learning (above) and learning-based (below) methods on four sequences from \textit{KITTI} dataset. The results of the compared methods are reported by BEVPlace~\cite{luo2023iccv}. The best result is in \textbf{bold} and the second best is \underline{underscored}.} 
  \resizebox{0.43\textwidth}{!}{
	\begin{tabular}{r|ccccc}
		\toprule
		Method & 00 & 02 & 05 & 06 & \textbf{Avg.}  \\
		\midrule
		Scan Context~\cite{kim2018iros}      & 89.7 & 73.9 & 77.0 & 86.7 & 81.8 \\
    BVMatch~\cite{luo2021ral}            & 93.8 & 78.2 & 90.2 & 93.8 & 89.0 \\
    \midrule
		PointNetVLAD~\cite{angelina2018cvpr} & 91.6 & 62.3 & 76.9 & 77.8 & 77.2 \\
		OverlapTransformer~\cite{ma2022ral}  & 96.7 & \underline{80.1} & 91.9 & \underline{95.6} & 91.1 \\
    BEVPlace~\cite{luo2023iccv} &\textbf{99.7} & \textbf{98.1} & \underline{99.3} &\textbf{100.0} & \textbf{99.3} \\
    \midrule
    Ours (local map context) & 96.0 & 73.0 & 98.3 & \textbf{100.0}  & 91.8 \\  
    Ours (full)              & \underline{99.4} & 78.4 & \textbf{100.0}   & \textbf{100.0}  & \underline{94.5} \\  
    \bottomrule
	\end{tabular}
  }
  \vspace{-8pt}
	\label{tab:loop_eval_kitti}
\end{table}

\begin{table*}[!t]
  \caption{3D reconstruction quality of different methods on \text{Quad} and \text{Math Institute} sequence of the \textit{Newer College} dataset~\cite{ramezani2020iros}. The reference model is measured by terrestrial laser scanning with mm-level accuracy. We report completion, accuracy and Chamfer-L1 in cm as well as F-score in $\%$ calculated with a 20\,cm error threshold. We highlight the best results in \textbf{bold} and the second best in \underline{underscored}. \\  - indicates the number is not reported and unavailable using the open-source code.}
  \setlength{\belowcaptionskip}{-6pt}
  \centering
     \resizebox{0.94\textwidth}{!}{
     \begin{tabular}{lc|cccc|cccc}
        \toprule
        \multirow{2}{*}{Method}& \multirow{2}{*}{Pose} & \multicolumn{4}{c|}{\text{Quad}}& \multicolumn{4}{c}{\text{Math Institute}}\\
        &&Map. Acc. $\downarrow$&Map. Comp. $\downarrow$&C-l1. $\downarrow$& F-score  $\uparrow $&Map. Acc. $\downarrow$&Map. Comp. $\downarrow$&C-l1. $\downarrow$& F-score $\uparrow $\\
        \toprule
        VDB-Fusion~\cite{vizzo2022sensors}& \multirow{3}{*}{KISS-ICP~\cite{vizzo2023ral}} & 14.03 &25.46&19.75& 69.50 & 15.21 & 28.66 & 21.94 & 63.35\\
        SHINE~\cite{zhong2023icra}& & 14.87
        & \underline{20.02} & \underline{17.45} & 68.85 & 14.46 & 34.03 & 24.24 & 64.38\\
        NKSR~\cite{huang2023cvpr} & & 15.67 & 36.87 & 26.27 & 58.57 & 15.11 & 27.10 & 21.11 & 65.08\\
        \midrule
        Puma~\cite{vizzo2021icra}&\multirow{4}{*}{Own Odometry} &  15.30&71.91& 43.60& 57.27 & 15.81 & 46.00 & 30.91 & 54.95 \\
        SLAMesh~\cite{ruan2023icra} & & 19.21 & 48.83 & 34.02 & 45.24 & \bf{12.80} & \underline{23.50} & \underline{18.16} & \underline{75.17} \\
        Nerf-LOAM~\cite{deng2023iccv} & & \underline{12.89} & 22.21 & 17.55 & \underline{74.37} & - & - & - & -\\
        PIN-SLAM && \bf{11.55} & \bf{15.25} & \bf{13.40} & \bf{82.08} & \underline{13.70} & \bf{21.91} & \bf{17.80} & \bf{75.49} \\
        \bottomrule
     \end{tabular}
     }
     \vspace{-0.1cm}
  \label{tab:recon_experiments_on_ncd}
\end{table*}

\begin{figure*}[t]
  \centering  
  \includegraphics[width=0.9\linewidth]{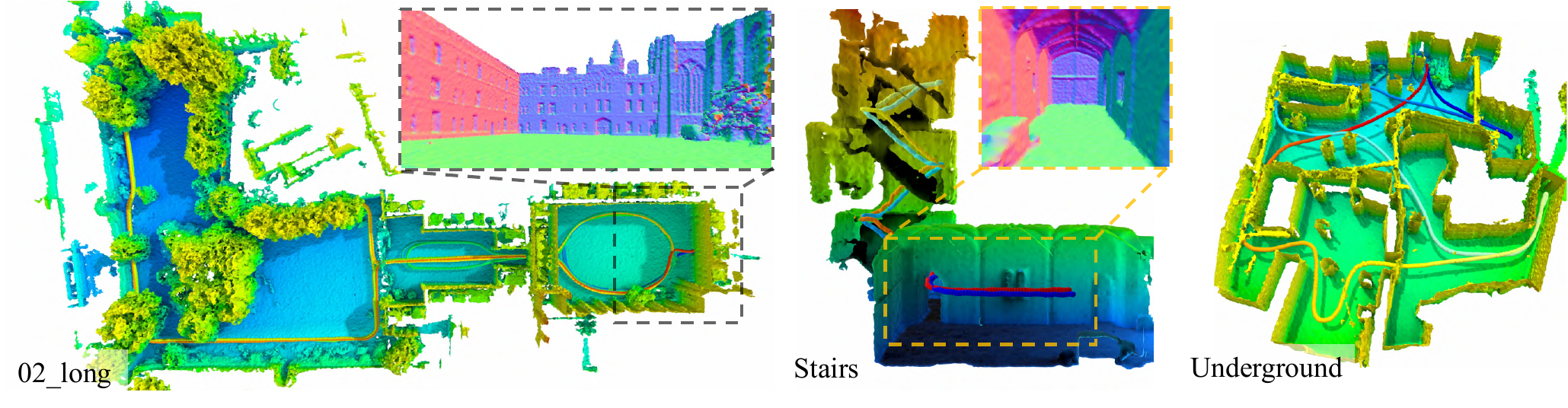}
  \caption{Globally consistent mesh reconstructed by PIN-SLAM on \textit{Newer College} dataset with multiple loops. The estimated trajectories are overlaid on the map and colorized according to the timestamp. Details of several revisited regions are highlighted in the dashed boxes.}
  \label{fig:ncd_quali}
  \vspace{-6pt}
\end{figure*}

\begin{figure*}[t]
  \centering  
  \includegraphics[width=0.92\linewidth]{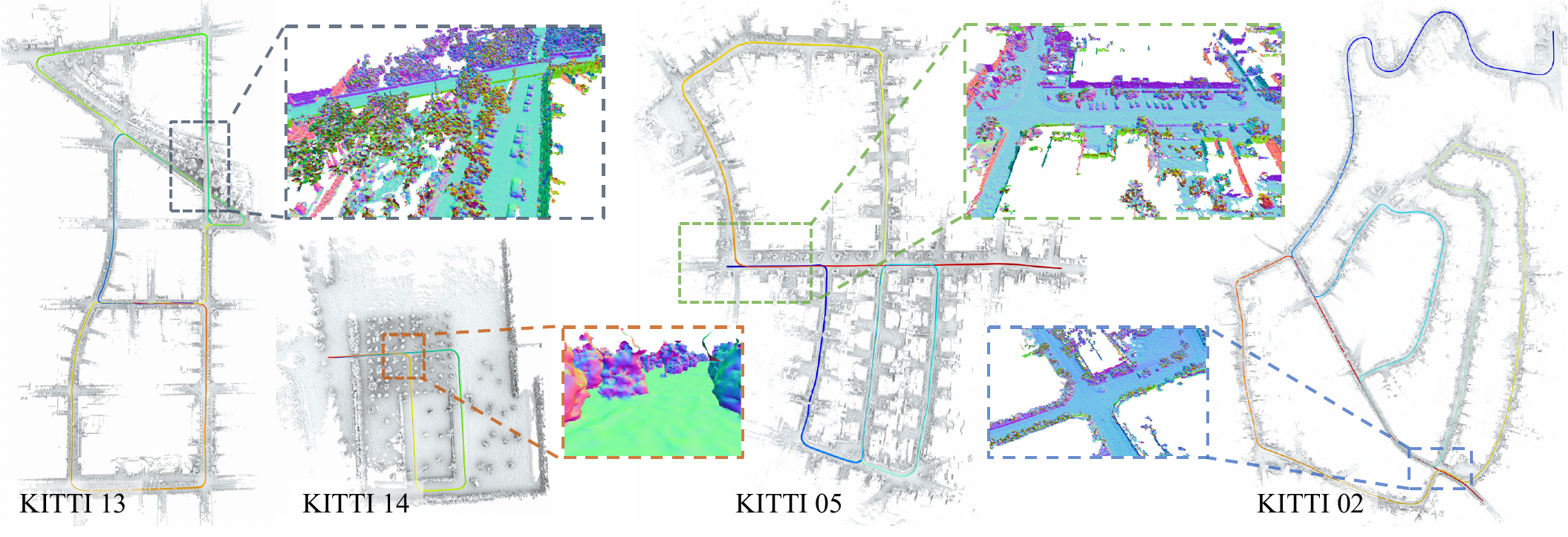}
  \caption{Globally consistent mesh reconstructed by PIN-SLAM on \textit{KITTI} dataset with multiple loops. The estimated trajectories are overlaid on the map and colorized according to the timestamp. Details of several revisited regions are highlighted in the dashed boxes.}
  \label{fig:kitti_quali}
  \vspace{-6pt}
\end{figure*}

\subsection{Evaluation of the Loop Closure Detection Performance}
\label{subsec:loopeval}

We follow BEVPlace~\cite{luo2023iccv} to evaluate loop closure detection performance on the KITTI dataset. We use the same partition of database and query frames as used in BEVPlace. We show the loop detection recall at Top-1 in \tabref{tab:loop_eval_kitti} and compare our approach with several non-learning and learning-based methods. Even without any offline pre-training, our approach achieves the second-best average recall after BEVPlace, outperforming the other compared methods. We show that using the polar context descriptor of neural points in the local map can improve the loop detection recall over the original scan context~\cite{kim2018iros}. Moreover, leveraging online optimized neural point features can enhance the distinctiveness of the context descriptor. Though our approach does not achieve the best performance in LiDAR place recognition, it proves to be adequate for a highly accurate SLAM system, without the need for pre-training or external map representations.

\subsection{Evaluation of the Mapping Performance}
\label{subsec:mappingeval}

\subsubsection{3D Reconstruction Quality}
In this section, we show the mapping quality of PIN-SLAM in terms of the 3D reconstruction quality of the resulting mesh in~\tabref{tab:recon_experiments_on_ncd}. 
The mesh is reconstructed from the SDF queried at the fixed-size grid using the marching cubes algorithm~\cite{lorensen1987siggraph}.
For the quantitative evaluation, we use the commonly used 3D reconstruction metrics~\cite{mescheder2019cvpr} calculated between the reconstructed and reference 3D model, namely accuracy, completeness, Chamfer-L1 distance, and F-score.
We select two scenes from the Newer College dataset~\cite{ramezani2020iros}, namely \text{Quad} from the \text{02\_long} sequence and \text{Math Institute} from the \text{math\_easy} sequence.
Both scenes have the survey-grade point cloud map measured by TLS available, which is taken as the reference model for evaluation. 
We compare PIN-SLAM against three state-of-the-art LiDAR SLAM approaches that can reconstruct dense 3D meshes, namely Puma~\cite{vizzo2021icra} based on Poisson surface reconstruction, SLAMesh~\cite{ruan2023icra} utilizing Gaussian process reconstruction, and Nerf-LOAM~\cite{deng2023iccv} employing implicit neural representation like us.
We additionally include three state-of-the-art ``mapping with known poses" methods tailored for LiDAR data, namely VDB-Fusion~\cite{vizzo2022sensors} based on TSDF integration, SHINE-Mapping~\cite{zhong2023icra} based on implicit neural scene reconstruction, and the data-driven method NKSR~\cite{huang2023cvpr} based on neural kernel field.
We use the pose estimation of KISS-ICP~\cite{vizzo2023ral} as the pose input to these three mapping methods. 
Note that KISS-ICP and PIN-SLAM achieve similar localization RMSE of about 10\,cm in both scenes. 
We use the same voxel size of 20\,cm for mesh reconstruction for the compared methods.
\tabref{tab:recon_experiments_on_ncd} lists the obtained results. 
As can be seen, PIN-SLAM achieves the best mapping quality in terms of completeness, Chamfer distance and F-score in both scenes, indicating that PIN-SLAM can achieve more accurate and more complete reconstruction of the environment than the compared methods. 
The superior performance of PIN-SLAM owes to the continuous implicit neural representation and the more consistent pose estimation.  

\begin{figure*}[t]
  \centering  
  \includegraphics[width=0.90\linewidth]{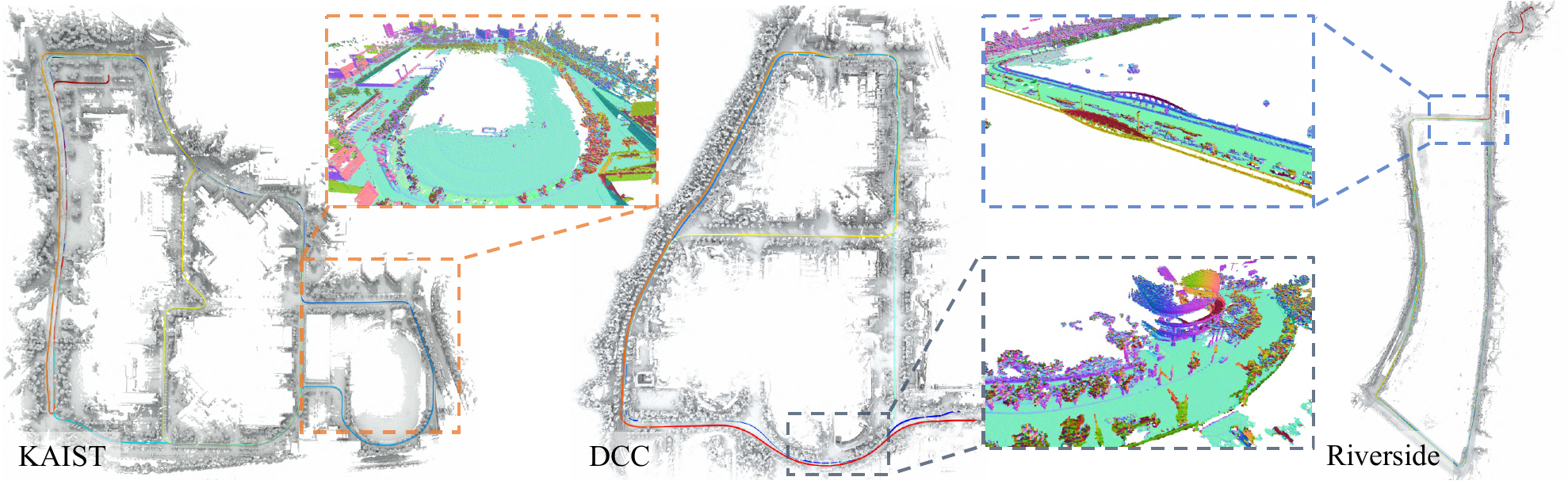}
  \caption{Globally consistent mesh reconstructed by PIN-SLAM on \textit{MulRAN} dataset with multiple loops. The estimated trajectories are overlaid on the map and colorized according to the timestamp. Details of several revisited regions are highlighted in the dashed boxes.}
  \label{fig:mulran_quali}
  \vspace{-10pt}
\end{figure*}

\begin{figure}[t]
  \centering  
  \includegraphics[width=0.90\linewidth]{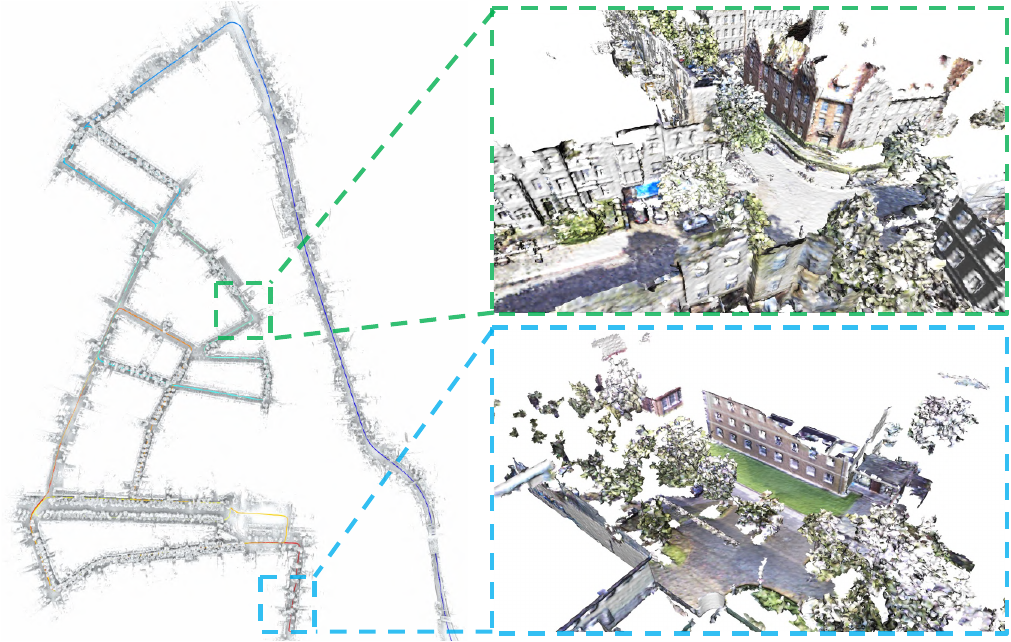}
  \setlength{\abovecaptionskip}{5pt}
  \caption{Qualitative results on the \textit{IPB-Car} dataset. On the left, we show the globally consistent mesh reconstructed by PIN-SLAM. On the right, we show two examples of the colorized mesh reconstructed using our calibrated LiDAR-camera multi-sensor platform.}
  \vspace{-8pt}
  \label{fig:ipbcar_quali}
\end{figure}

\subsubsection{Map Consistency}
Since the global localization accuracy can reflect the mapping consistency, we qualitatively depict the globally consistent mesh reconstructed from PIN map built by PIN-SLAM on Newer College in \figref{fig:ncd_quali}, KITTI in \figref{fig:kitti_quali}, MulRAN in \figref{fig:mulran_quali}, and IPB-Car dataset in \figref{fig:ipbcar_quali}.
Several regions with multiple loop closures are highlighted.

\subsection{Ablation studies}

In this subsection, we conduct ablation studies to validate our design choices.
We use the average relative translational drifting error (ARTE) on KITTI dataset as the metrics and the same random seed to carry out the studies. 
We show ARTE obtained on KITTI sequences \text{00-10} with different design choices for mapping and odometry in \tabref{tab:ablation_losses} and \tabref{tab:ablation_odom}, respectively.
%
%
For the mapping losses, we find out that the BCE loss used by our method is more suitable for SDF training than L1 or L2 loss. 
Besides, Ekional loss is necessary for the odometry and a weight $\lambda_e$ of $0.5$ is preferable. 
To calculate the Ekional loss, using the numerical gradient rather than the analytical gradient leads to a smooth and regular SDF in the free space, thus improving localization accuracy.
A perturbation step $\epsilon$ of 8\,cm for numerical gradient calculation is preferable.
For odometry estimation, we verify that the usage of GM robust kernel for SDF residual and SDF gradient anomaly are both beneficial to localization accuracy. 
The odometry loses track when using the numerical gradient because it is often less accurate than the analytical one once the SDF has been fitted.
Besides, as shown in \tabref{tab:ablation_resolution}, we find out that a too-large or too-small neural point resolution $v_p$ would lead to suboptimal odometry accuracy. 
Setting $v_p$ as 0.4\,\si{\metre}, corresponding to our adaptive setting $0.005 r_{\text{max}}$, leads to the smallest ARTE on all three tested sequences. 

\begin{table}[t]
  \centering
  \caption{Ablation study: design choices of map training losses. ARTE represents the average relative translational drifting errors on KITTI sequence \text{00-10}. \xmark ~denotes failure.}
  \resizebox{0.38\textwidth}{!}{
  \begin{tabular}{cccc|c}
    \toprule
    SDF loss & Ekional loss $\lambda_{e}$  & num. grad. & $\epsilon$ [\si{\metre}]&ARTE [\%] $\downarrow$  \\
    \midrule
     BCE         & 0.5     & \xmark        &  -      &  0.57        \\
     BCE        & 1.0     & \cmark        & 0.08      &  0.59  \\
     BCE     & 0.2  & \cmark      & 0.08      &  0.55    \\
     BCE    & 0.0    & -            & -     &  \xmark  \\
     L1      & 0.5   & \cmark       & 0.08       &    1.15   \\
     L2   & 0.5   & \cmark       & 0.08     & \xmark     \\
     BCE   & 0.5    & \cmark & 0.04   &   0.54            \\
     BCE   & 0.5    & \cmark & 0.16   &  1.48          \\
    \midrule
     BCE   & 0.5    & \cmark & 0.08    &  \bf{0.50}     \\
    \bottomrule
  \end{tabular}
  }
  \label{tab:ablation_losses}
  \vspace{-4pt}
\end{table}

\begin{table}[htbp]
  \centering
  \caption{Ablation study: design choices of odometry estimation. ARTE represents the average relative translational drifting errors on KITTI sequence \text{00-10}. \xmark ~denotes failure.}
  \resizebox{0.36\textwidth}{!}{
  \begin{tabular}{ccc|c}
    \toprule
    GM kernel $w_{r}$ & GM kernel $w_{g}$  & anl. grad.  &  ARTE [\%] $\downarrow$   \\
    \midrule
    \xmark    & \xmark    & \cmark & 0.56      \\
    \cmark    & \xmark    & \cmark & 0.52      \\
    \xmark    & \cmark    & \cmark & 0.55      \\
    \cmark    & \cmark    & \xmark & \xmark     \\
    \midrule
    \cmark    & \cmark    & \cmark & \bf{0.50}  \\
    \bottomrule
  \end{tabular}
  }
  \label{tab:ablation_odom}
  \vspace{-6pt}
\end{table}

\begin{table}[!t]
  \centering
  \caption{Ablation study: average relative translational drifting error [\%] with regards to different neural point resolutions $v_p$.}
  \resizebox{0.35\textwidth}{!}{
  \begin{tabular}{c|ccccc}
  \toprule
  \multirow{2}{*}{Sequence} & \multicolumn{5}{c}{Neural point resolution ~$v_p$}                                       \\
	& {1.0\,\si{\metre}}  & 0.8\,\si{\metre} & 0.6\,\si{\metre} & 0.4\,\si{\metre} & 0.2\,\si{\metre}     \\
  \midrule
  KITTI \text{00} & 0.98  & 0.75 & 0.85 & \bf{0.55} & 1.06 \\
  KITTI \text{05} & 0.36  & 0.56 & 0.45 & \bf{0.29} & 0.50 \\
  KITTI \text{08} & 0.96  & 0.88 & 0.85 & \bf{0.83} & 1.05 \\
  \bottomrule
  \end{tabular}
  }
  \vspace{-6pt}
  \label{tab:ablation_resolution}
\end{table}

\begin{table}[!t]
  \centering
  \caption{Memory consumption in MB and the compression ratio with regards to the raw point cloud of different map representations.}
  \resizebox{0.48\textwidth}{!}{
  \begin{tabular}{ccccc}
  \toprule
  Representation & KITTI \text{00} & KITTI \text{05} & KITTI \text{08} & NCD \text{02} \\
  \midrule
  Raw point cloud  & 13624.2 & 8284.7 & 12214.1 & 26559.0 \\
  Surfel map & 887.7 & 512.6 & 835.7 & 79.0 \\
  Mesh map & 2032.9 & 1317.4 & 1894.1 & 1503.7 \\
  VDB TSDF map & 748.1 & 434.6 & 958.6 & 462.5 \\ 
  SHINE map & 160.6 &  114.2 & 189.9 & 117.1 \\ 
  \midrule
  PIN-LO map  & 137.3 & 89.6 & 162.4  & 110.3 \\
  PIN-SLAM map & \bf{102.1 (0.7\%)} & \bf{66.3 (0.8\%)} & \bf{138.8 (1.1\%)}  & \bf{76.8 (0.3\%)} \\
  \bottomrule
  \end{tabular}
  }
  \label{tab:mapmemory}
  \vspace{-8pt}
\end{table}

\subsection{Memory and Computational Resources}

\subsubsection{Memory Requirement}
We report the map memory consumption of PIN-SLAM on four representative sequences from KITTI and Newer College dataset in \tabref{tab:mapmemory}. 
We compare the proposed point-based implicit neural (PIN) map with several common map representations: surfel map used by SuMa~\cite{behley2018rss}, mesh map used by Puma~\cite{vizzo2021icra}, VDB TSDF map used by VDB Fusion~\cite{vizzo2022sensors}, and the grid-based sparse hierarchical implicit neural (SHINE) map used by SHINE-Mapping~\cite{zhong2023icra} and Nerf-LOAM~\cite{deng2023iccv}.
We also provide the raw point cloud memory consumption as a reference. 
For the VDB TSDF map, we use a voxel size of 20\,cm.
For SHINE map and PIN map, we use the same resolution of 40\,cm for the local latent feature grid and neural points. With the continuous neural SDF queried from SHINE or PIN map, we can reconstruct the mesh with a voxel size of 20\,cm with a similar quality as the mesh reconstructed from the discrete VDB TSDF map.
Besides, for VDB TSDF map and SHINE map, we take the odometry estimation of KISS-ICP~\cite{vizzo2023ral} as their pose.
As shown in \tabref{tab:mapmemory}, our PIN map is the most compact representation among the compared methods with a compression ratio of about 0.7\% on average.
Compared to SHINE map, PIN map using the same resolution and feature dimensions is about 15\% more memory efficient even with the additional storage of neural point coordinates and orientations. 
This is because PIN map does not have a hierarchical tree structure like SHINE map and the neural points are only allocated close to the surface.         
Notably, compared with PIN LiDAR odometry, the memory consumption of PIN map decreases by about 20\% due to the loop correction and map update of PIN-SLAM.
The elastic PIN map can deform with the corrected pose and eliminate duplicated neural points representing the same place.
This is not possible for the grid-based VDB or SHINE map without the submap scheme or offline remapping.

\subsubsection{Computational Requirement}
\begin{figure}[!t]
  \centering  
  \includegraphics[width=\linewidth]{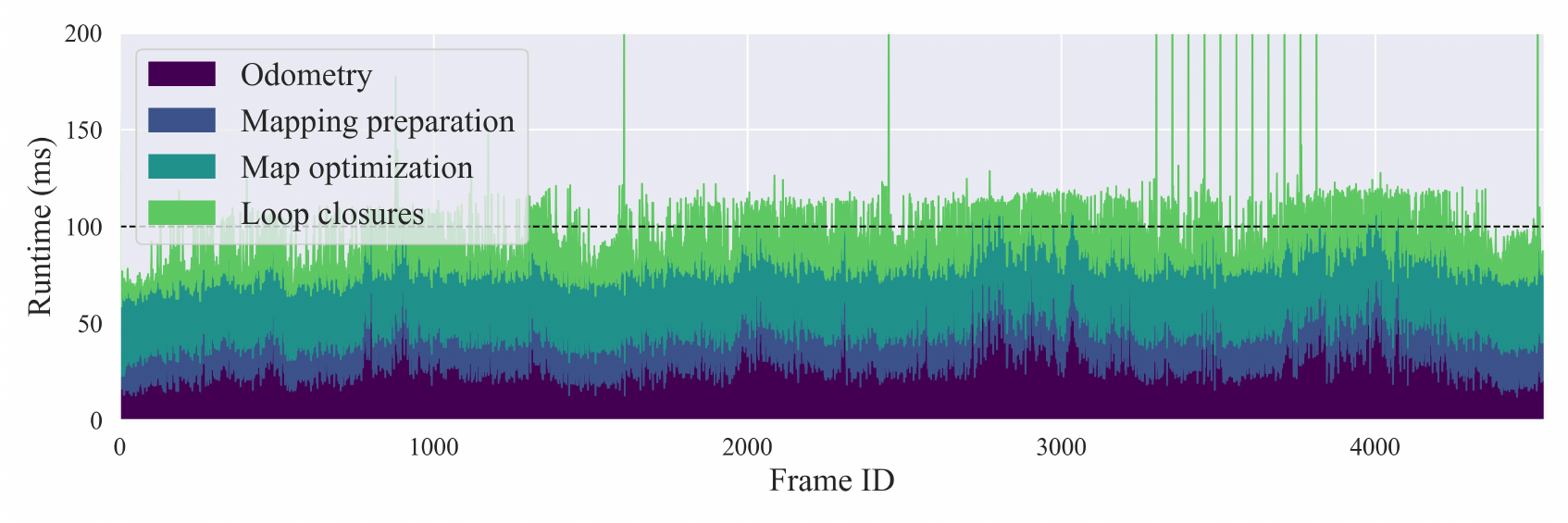}
  \setlength{\abovecaptionskip}{-12pt}
  \caption{Processing time for each sub-module of the light version of PIN-SLAM on KITTI sequence 00.}
  \label{fig:timing}
  \vspace{-3pt}
\end{figure}

In \tabref{tab:time_ate}, we report the average runtime of PIN-SLAM running on a single NVIDIA Quadro A4000 GPU over the whole 11 sequences of the KITTI dataset. 
We show the results of two versions of PIN-SLAM, the full version as used in the localization evaluation above and the light version which takes a fewer number of iterations during mapping and odometry.
The light version of PIN-SLAM works 38\% faster at the cost of a minor degradation (10\%) of the localization accuracy. 
It can operate efficiently at approximately 11\,Hz, aligning with the typical sensor frame rate. 
In~\figref{fig:timing}, we report the average processing time for each sub-module of PIN-SLAM (light).
%
%
Odometry and map optimization are the two most time-consuming parts, each taking up about 40\% of the total run time.
When loops are detected and pose graph optimization is conducted, the run time per frame occasionally exceeds 200\,ms (as the spikes in the figure).
Notably, the processing time remains consistent as the frame count increases.
In contrast, the only implicit neural LiDAR odometry baseline, Nerf-LOAM~\cite{deng2023iccv} takes more than 4 seconds per frame, which is 30$\times$ slower than PIN-SLAM.
Moreover, we observe that Nerf-LOAM's processing time increases drastically with an increasing number of scans.

\begin{table}[!t]
  \centering
  \caption{Comparison of the average operation speed and the localization error on KITTI sequence 00-10 using a single NVIDIA A4000 GPU. ARTE represents the average relative translational error.}
  \resizebox{0.44\textwidth}{!}{
  \begin{tabular}{c|ccc}
  \toprule
  Method  & Time per frame [s] $\downarrow$ & FPS [Hz] $\uparrow$ & ARTE [\%] $\downarrow$  \\
  \midrule
  PIN-SLAM (full)  & 0.14  & 7.1 & \bf{0.50}  \\
  PIN-SLAM (light) & \bf{0.09} & \bf{11.3} & 0.56 \\
  Nerf-LOAM \cite{deng2023iccv} & 4.43 & 0.2  & 1.69 \\
  \bottomrule
  \end{tabular}
  }
  \vspace{-3pt}
  \label{tab:time_ate}
\end{table}

\subsection{Extension on Semantic Mapping}

\begin{figure}[!t]
  \centering  
  \includegraphics[width=0.97\linewidth]{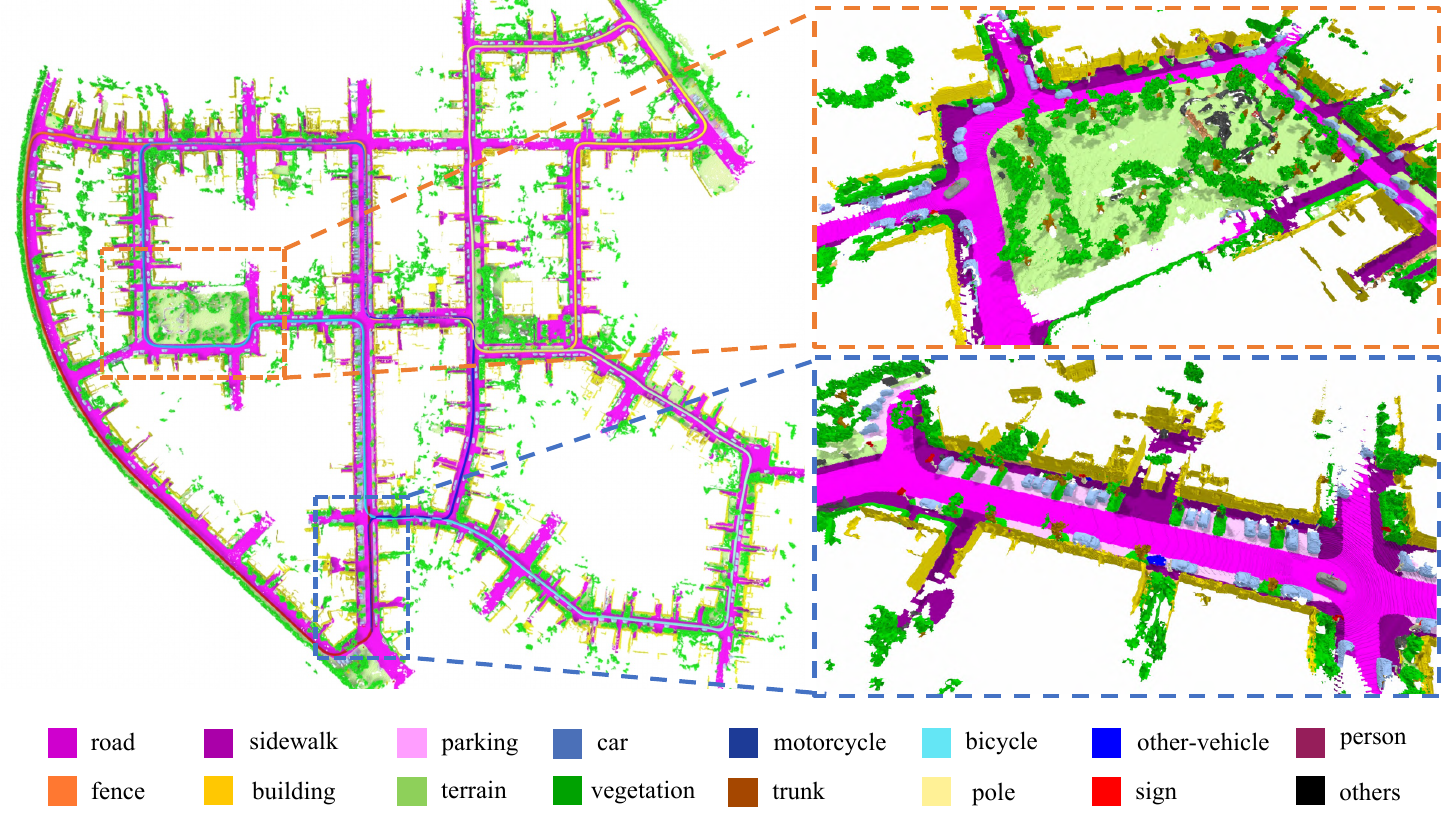}
  \setlength{\abovecaptionskip}{3pt}
  \caption{An example of the metric-semantic map built by PIN-SLAM on KITTI sequence \text{00} using the semantic labels of SemanticKITTI.}
  \label{fig:semantic_slam}
  \vspace{-8pt}
\end{figure}

As a proof of concept, we illustrate briefly that our approach is capable of conducting metric-semantic SLAM with global consistency.
As shown in \figref{fig:semantic_slam}, PIN-SLAM manages to build a metric-semantic map using the SemanticKITTI~\cite{behley2019iccv} dataset by integrating the semantic labels as described in \secref{sec:extension}.
Note that we only need to add another shallow MLP for the semantic querying.
Currently, the semantics do not play a role in odometry estimation and loop closure detection.
Considering this is outside the scope of this work but a possible future extension.

\section{Conclusion}
\label{sec:conclusion}

In this paper, we presented PIN-SLAM, a novel LiDAR SLAM approach to perform globally consistent mapping using a point-based implicit neural map representation.
%
%
Our approach alternates between the online incremental learning of the local implicit map and the odometry estimation using a correspondence-free point-to-implicit map registration. 
We exploit sparse neural points as local feature embeddings, which are inherently elastic and deformable throughout the global pose adjustment when correcting a loop closure. 
This enables us to effectively maintain the global consistency of both the neural points and the underlying implicit map.
We implemented and evaluated our approach on various datasets, provided comparisons to other existing techniques and supported
all claims made in this paper. 
Extensive experiments suggest PIN-SLAM achieves better or on-par localization accuracy than previous methods.  
Additionally, it builds a more consistent and compact implicit map, which can be reconstructed into more accurate and complete meshes.
Besides, PIN-SLAM can run at the sensor frame rate using a moderate GPU.

\textbf{Limitations and Future Work.} 
Future work may use additionally an IMU for a more robust, accurate and efficient SLAM system. 
%
%
Besides, we are currently using a fixed resolution of neural points.
%
We can enhance the reconstruction quality by adaptively distributing and dynamically moving neural points throughout the scene like previous works~\cite{sandstrom2023iccv, li2022cvpr-dccdif}, and use a data structure such as i-Kdtree~\cite{xu2022tro} enabling efficient neighbor search without using voxel structures. 
Another direction can be using semantic information to further enhance odometry estimation and loop closure detection.



\section{Acknowledgements}
\label{sec:wcknowledgements}
We thank Ignacio Vizzo, Benedikt Mersch, Yibin Wu, Tiziano Guadagnino, Haofei Kuang for the fruitful discussions and Rodrigo Marcuzzi for the colorized point cloud. 


\bibliographystyle{plain_abbrv}
\bibliography{glorified,new}

\vspace{-8mm}
\begin{IEEEbiography}[{\includegraphics[width=1in,height=1.25in,clip,keepaspectratio]{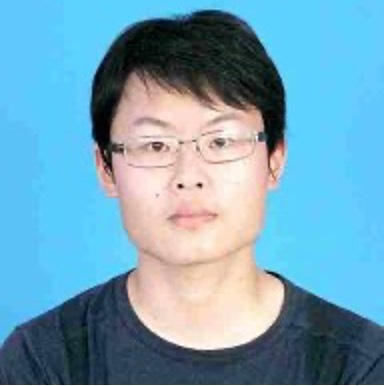}}]{Yue Pan} is a Ph.D. student in Engineering at the Photogrammetry \& Robotics Lab headed by Prof. Cyrill Stachniss at the University of Bonn, Germany. He obtained his B.Sc. degree in Geomatics Engineering from Wuhan University, China in 2019 and received his MSc degree in Geomatics Engineering from ETH Zurich, Switzerland in 2022. His research focuses on  SLAM, 3D reconstruction and navigation.
\end{IEEEbiography}

\vspace{-15mm}
\begin{IEEEbiography}[{\includegraphics[width=1in,height=1.25in,clip,keepaspectratio]{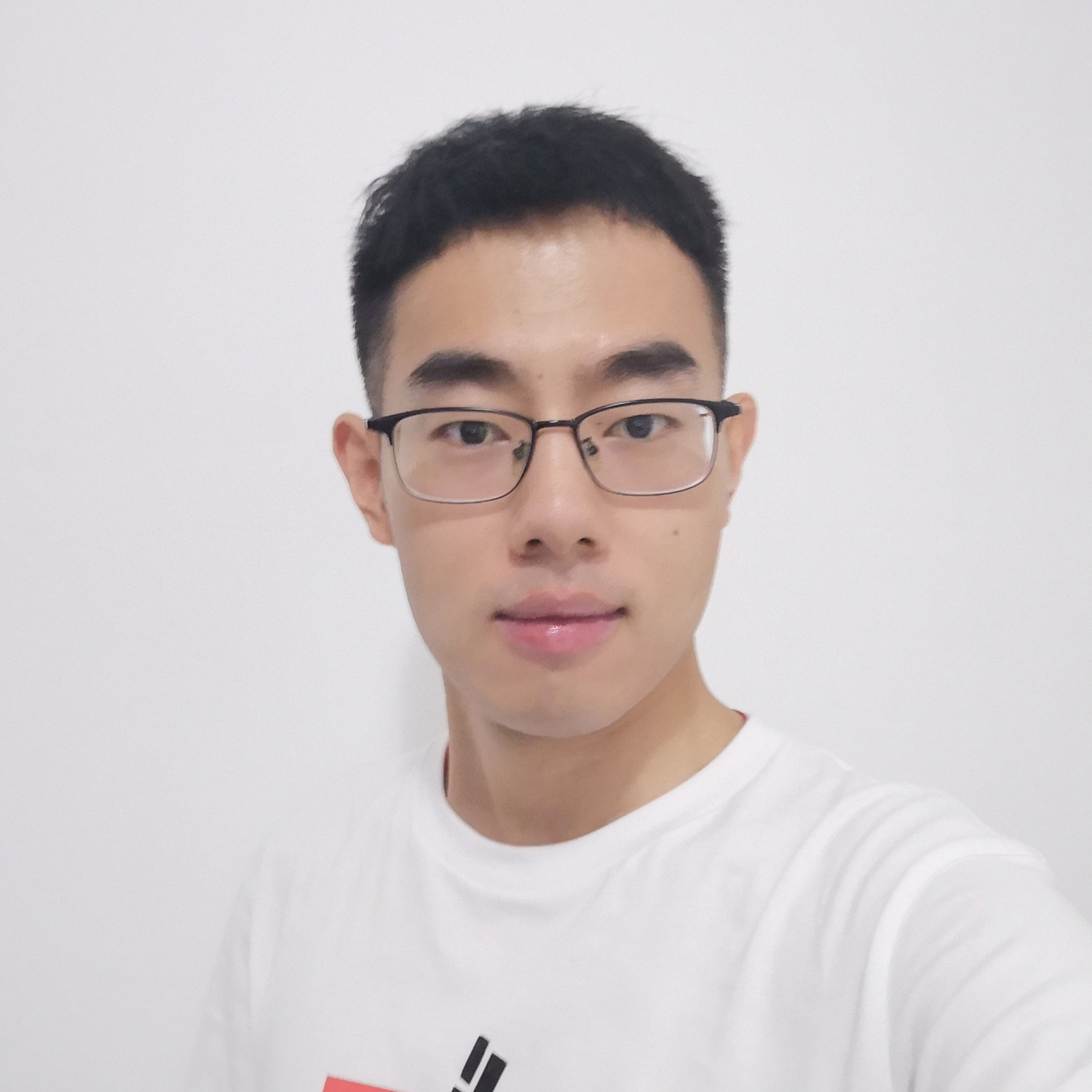}}]{Xingguang Zhong} is a Ph.D. student in Engineering at the Photogrammetry \& Robotics Lab at the University of Bonn, Germany headed by Prof. Cyrill Stachniss. He obtained his B.Sc. degree in Mechanical Engineering in 2017 and his M.Sc. degree in Mechatronic Engineering in 2019 from Harbin Institute of Technology, China. His research interests include large-scale 3D reconstruction and autonomous navigation.
\end{IEEEbiography}

\vspace{-15mm}
\begin{IEEEbiography}[{\includegraphics[width=1in,height=1.25in,clip,keepaspectratio]{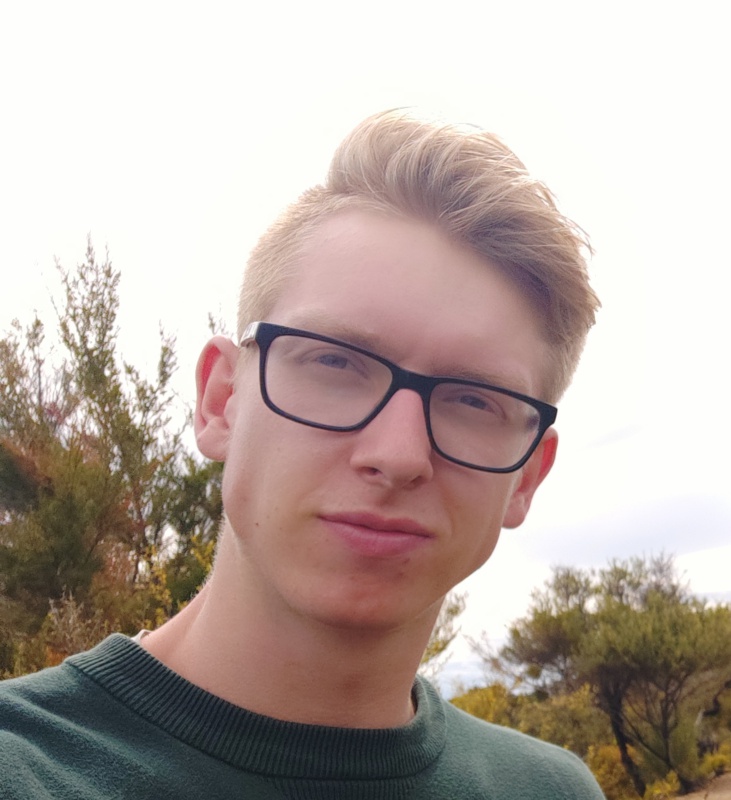}}]{Louis Wiesmann}  is a Ph.D. student at the Photogrammetry \& Robotics Lab headed by Prof. Cyrill Stachniss at the University of Bonn, Germany. He obtained his B.Sc. degree in 2017 and his M.Sc. degree in 2019 in Geodetic Engineering from the University of Bonn. His research focuses on localization and mapping in large-scale environments.
\end{IEEEbiography}

\vspace{-15mm}
\begin{IEEEbiography}[{\includegraphics[width=1in,height=1.25in,clip,keepaspectratio]{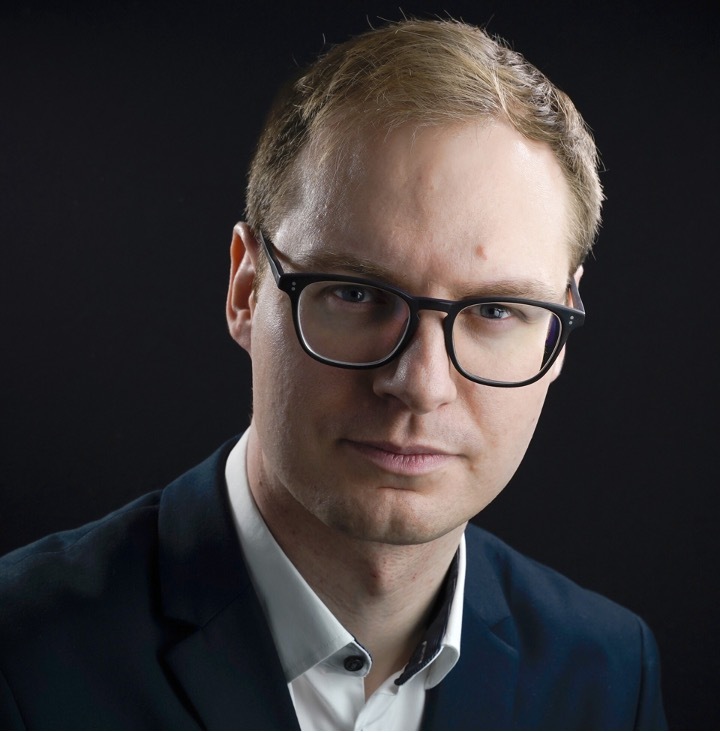}}]{Thorbjörn Posewsky} is a Ph.D. student in Engineering at the Photogrammetry \& Robotics Lab headed by Prof. Cyrill Stachniss at the University of Bonn, Germany and additionally software engineer at MicroVision GmbH, Hamburg, Germany. He obtained his B. Sc. degree in computer science in 2012 and his M. Sc. also in computer science in 2015, both from the University of Paderborn, Germany. His research focuses on Mapping and SLAM.
\end{IEEEbiography}

\vspace{-15mm}
\begin{IEEEbiography}[{\includegraphics[width=1in,height=1.25in,clip,keepaspectratio]{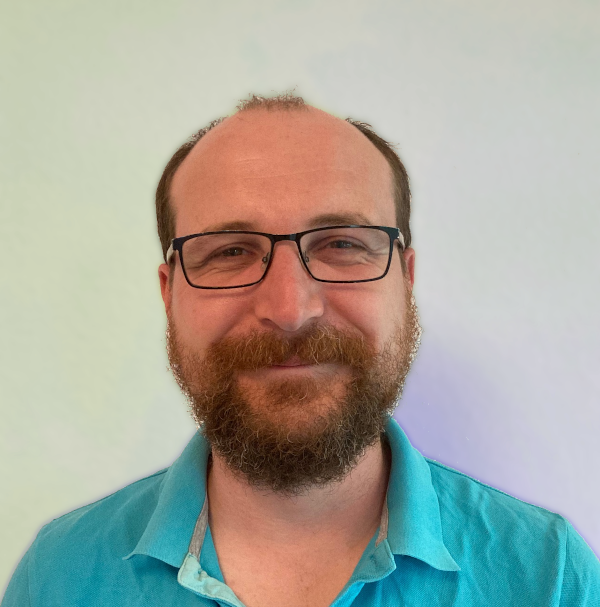}}]{Jens Behley} received his Dipl.-Inform.  in computer science in 2009 and his Ph.D. in computer science in  2014, both from the Dept. of Computer Science at the University of Bonn, Germany. Since 2016, he is a postdoctoral researcher at the Photogrammetry \& Robotics Lab at the University of Bonn, Germany. He finished his habilitation at the University of Bonn in 2023. His area of interest lies in the area of perception for autonomous vehicles, deep learning for semantic interpretation, and LiDAR-based SLAM.
\end{IEEEbiography}

\vspace{-15mm}
\begin{IEEEbiography}[{\includegraphics[width=1in,height=1.25in,clip,keepaspectratio]{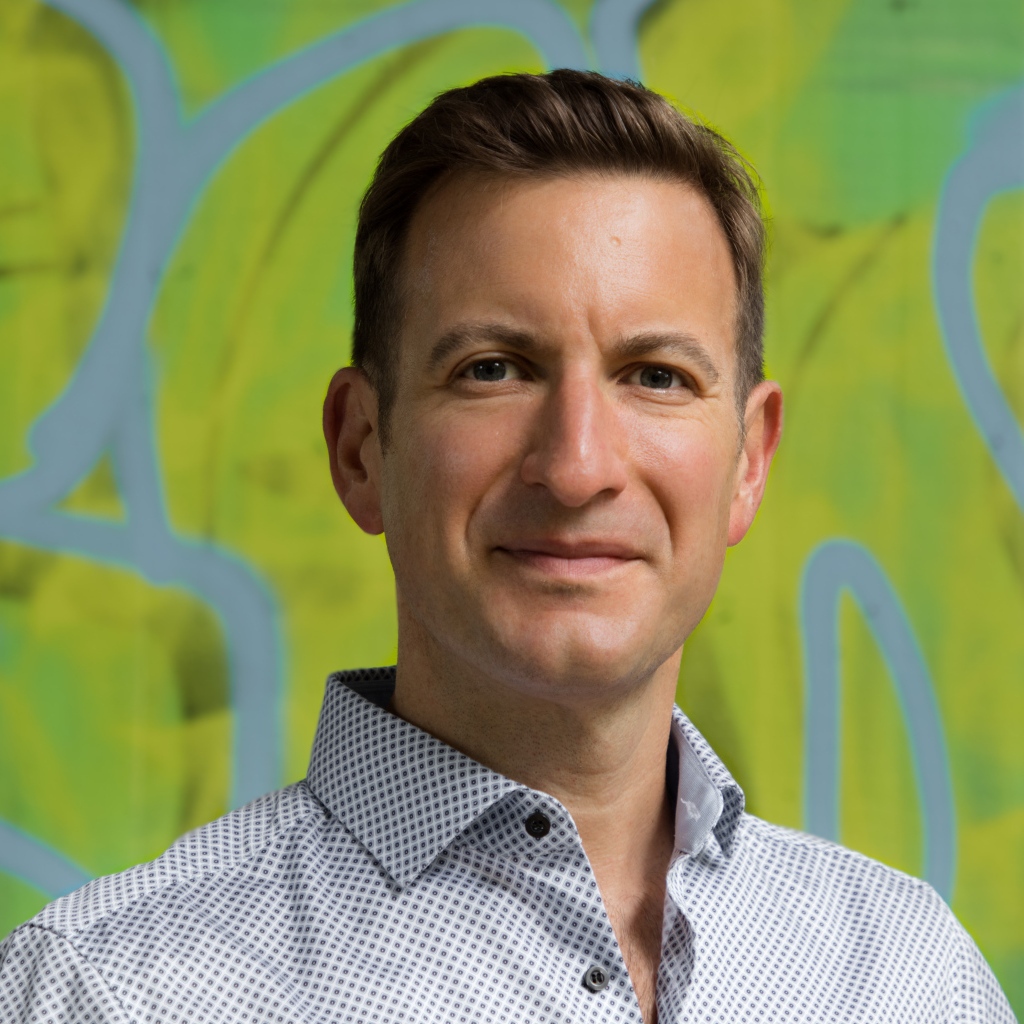}}]{Cyrill Stachniss} is a full professor at the University of Bonn, Germany, with the University of Oxford, UK, as well as with the Lamarr Institute for Machine Learning and AI, Germany. He is the Spokesperson of the DFG Cluster of Excellence PhenoRob at the University of Bonn. His research focuses on probabilistic techniques and learning approaches for mobile robotics, perception, and navigation. The main application areas of his research are agricultural robotics, service robotics, and self-driving cars.
\end{IEEEbiography}

\end{document}